\documentclass[journal, final]{IEEEtran}
\usepackage{lineno}

\usepackage{cite}

\ifCLASSINFOpdf
  \usepackage[pdftex]{graphicx}
\else
  \usepackage[dvips]{graphicx}
\fi
\usepackage{amsmath}
\usepackage{amssymb}

\usepackage{multirow}
\usepackage{capt-of}
\usepackage{xcolor}

\begin{document}
%
\title{Zero Shot Detection}
%
%
%

\author{Pengkai~Zhu	,~\IEEEmembership{Student~Member,~IEEE,}
        Hanxiao~Wang,~\IEEEmembership{Member,~IEEE,}
      and~Venkatesh~Saligrama,~\IEEEmembership{Fellow,~IEEE}
\thanks{The authors are with the Department of Electrical and Computer Engineering, Boston University, Boston, Massachusetts 02215, USA.(e-mail:
{zpk, hxw, srv}@bu.edu)}%
\thanks{Copyright \textcopyright \ 2019 IEEE. Personal use of this material is permitted. However, permission to use this material for any other purposes must be obtained from the IEEE by sending an email to pubs-permissions@ieee.org.}}

%
%

\markboth{IEEE TRANSACTIONS ON CIRCUITS AND SYSTEMS FOR VIDEO TECHNOLOGY}%
{Zhu \MakeLowercase{\textit{et al.}}: Zero Shot Detection}
%



\maketitle

\begin{abstract}
As we move towards large-scale object detection, it is unrealistic to expect annotated training data, in the form of bounding box annotations around objects, for all object classes at sufficient scale, and so methods capable of unseen object detection are required. We propose a novel zero-shot method based on training an end-to-end model that fuses  semantic attribute prediction with visual features to propose object bounding boxes for seen and unseen classes. While we utilize semantic features during training, our method is agnostic to semantic information for unseen classes at test-time. Our method retains the efficiency and effectiveness of YOLOv2~\cite{redmon2016yolo9000} for objects seen during training, while improving its performance for novel and unseen objects. The ability of state-of-art detection methods to learn discriminative object features to reject background proposals also limits their performance for unseen objects.

We posit that, to detect unseen objects, we must incorporate  semantic information into the visual domain so that the learned visual features reflect this information and leads to improved recall rates for unseen objects. We test our method on {\em PASCAL VOC} and {\em MS COCO} dataset and observed significant improvements on the average precision of {\em unseen} classes.   
\end{abstract}


%
\IEEEpeerreviewmaketitle

\section{Introduction}





\IEEEPARstart{Z}{ero} shot learning (ZSL) has recently drawn increasing attention\cite{zhang2015zero, fu2017recent, zhang2016zero, ECCVzhang2016zero, zhang2016zeroeccv}. By leveraging the inexpensive descriptions of categories, connections between the visual and semantic representations are built to interpolate the unseen class, which has no examples in the training stage. While methods like \cite{zhang2016zero} achieve high classification accuracy over unseen categories, a more realistic problem, generalized zero shot learning (gZSL), where both seen and unseen classes need to be recognized, is proposed\cite{xian2017zero}. This problem makes ZSL more practical since there is no oracle in the real world indicating whether an object class has been seen during training.

However, this extension of the classical ZSL problem still has its own limitation: it assumes the object is precisely localized and the only task is to recognize it. In fact, there are many potentially unseen objects appearing in the wild. An intelligent system should be able to not only classify, but also to localize them. Therefore, in this paper, we consider this additional source of complexity, and introduce zero shot detection (ZSD). This involves detecting unseen objects.

Deep learning based object detection methods trained on fully annotated training data has had significant success over the last few years~\cite{girshick2015fast,ren2015faster,redmon2016you,liu2016ssd,he2017mask,redmon2016yolo9000}. Methods such as YOLOv2~\cite{redmon2016yolo9000}, achieve high-performance, by training detectors to accurately predict object bounding boxes to match the ground-truth bounding boxes, while suppressing bounding boxes that are part of the background image.

Nevertheless, YOLOv2 as currently structured, is also somewhat limited in scope in the context of large-scale applications where we encounter a large number of object classes. In these applications, it is unrealistic to expect object bounding boxes for all classes at sufficient scale that is required for training. Indeed, the lack of labeled training data in large-scale recognition applications has led to the emergence of zero-shot recognition (ZSR) methods (see for instance, \cite{lampert2014attribute,lei2015predicting}) as an alternative means to supplement labeled data. As object detection moves towards large-scale\footnote{ 
Although the number of annotations have increased in common detection datasets (e.g. 20 classes provided by {\it PASCAL VOC}~\cite{everingham2010pascal} and 80 provided by the more recent {\it MSCOCO}~\cite{lin2014microsoft}), it is substantially smaller relative to image classification~\cite{deng2009imagenet}.}, it is imperative that we move towards a framework that serves the dual role of detecting objects seen during training as well as detecting heretofore unseen categories. %
Furthermore, this problem gains more importance as we move towards object detection appearing in the wild.%

Motivated by these challenges, we develop a novel zero-shot detection architecture (ZS-YOLO) for detection of unseen object classes. Our method is based on a seamless integration of semantic attribute predictors with YOLOv2's visual detection architecture\footnote{We choose YOLOv2 for concreteness. Our method is applicable to other architectures. See Sec.III.B}. Specifically, we train an end-to-end model for zero-shot detection based on a novel multi-task loss objective, which incorporates semantic and visual information. Nevertheless, at test-time, our method is agnostic to  semantic information of unseen objects, and the semantic component of our network functions as a system for identifying semantic components that resemble trained classes. We choose YOLOv2 as the base detector for zero-shot detection because it is the state-of-the-art single stage detector on existing benchmark datasets\cite{redmon2016yolo9000}. By changing the confidence loss and network backbone, our method can be easily applied to other single stage detector like SSD\cite{liu2016ssd} and RetinaNet\cite{lin2017focal}. In addition, ZS-YOLO can be viewed as a variation of region proposal network (RPN), thus can be integrated with two-stage detectors like Faster-RCNN\cite{ren2015faster} seamlessly. Ultimately, our choice of YOLOv2 is coincidental based partially on the ease with which we can integrate other side information, and the fundamental focus of the paper is on understanding and quantifying the utility of semantic attributes for zero-shot detection.

{\it Limitations of Naive Methods:} In comparison various naive strategies do not perform well. For instance, a cascade of YOLOv2 and a off-the-shelf zero-shot classifier (ZSR) at run-time turns out to be somewhat less effective in detecting unseen classes. This may seem surprising particularly because ZSR is capable of transferring semantic class-level information (such as attributes~\cite{farhadi2009describing,lampert2014attribute,mensink2012metric,parikh2011interactively, castanon2016retrieval, chen2018probabilistic}, word phrases~\cite{socher2013zero,frome2013devise}) for synthesizing unseen object classifiers.  

\begin{figure}[t]
\centering
\begin{tabular}{ll}
\includegraphics[width=0.45\linewidth]{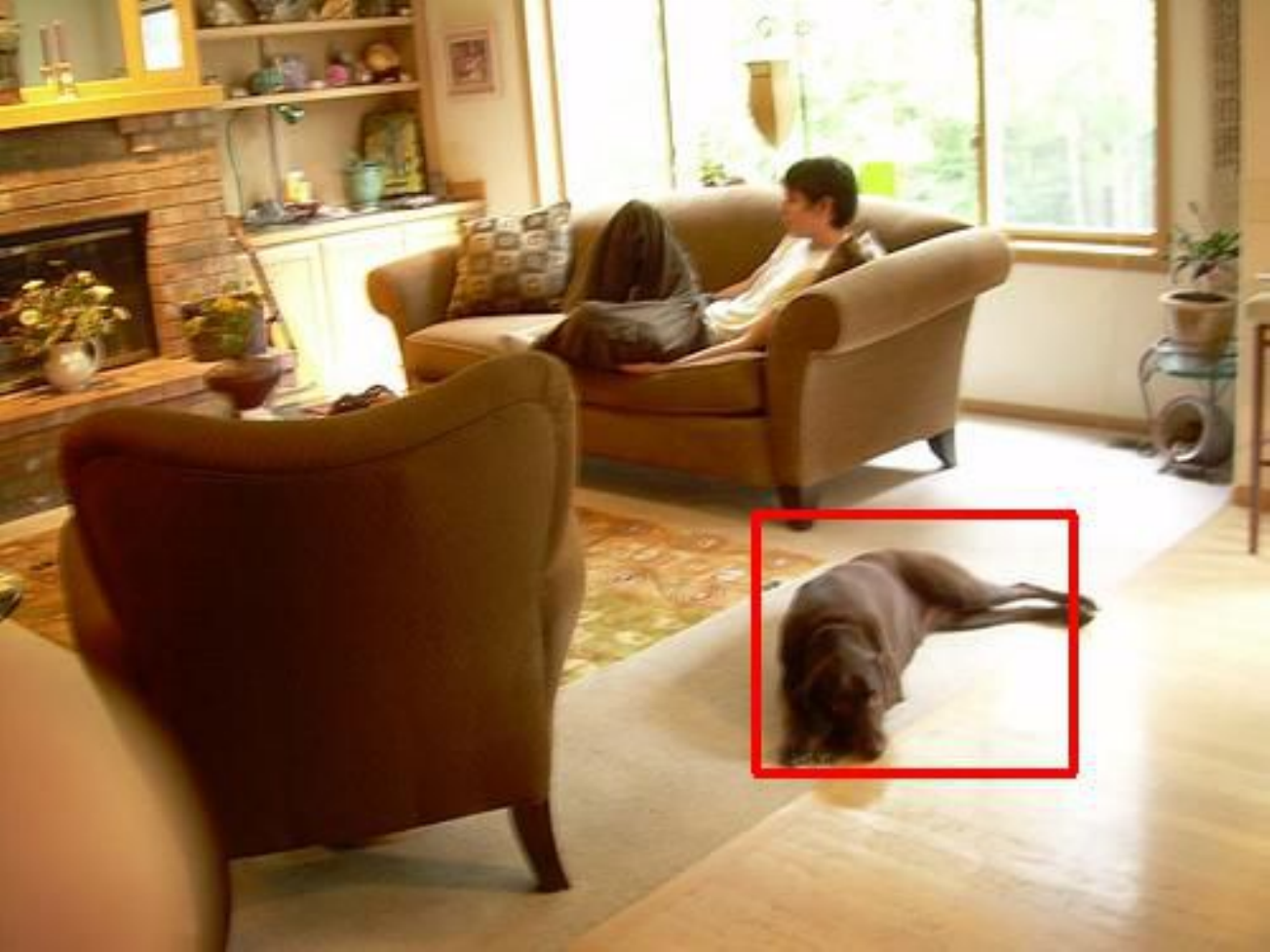}&
\includegraphics[width=0.45\linewidth]{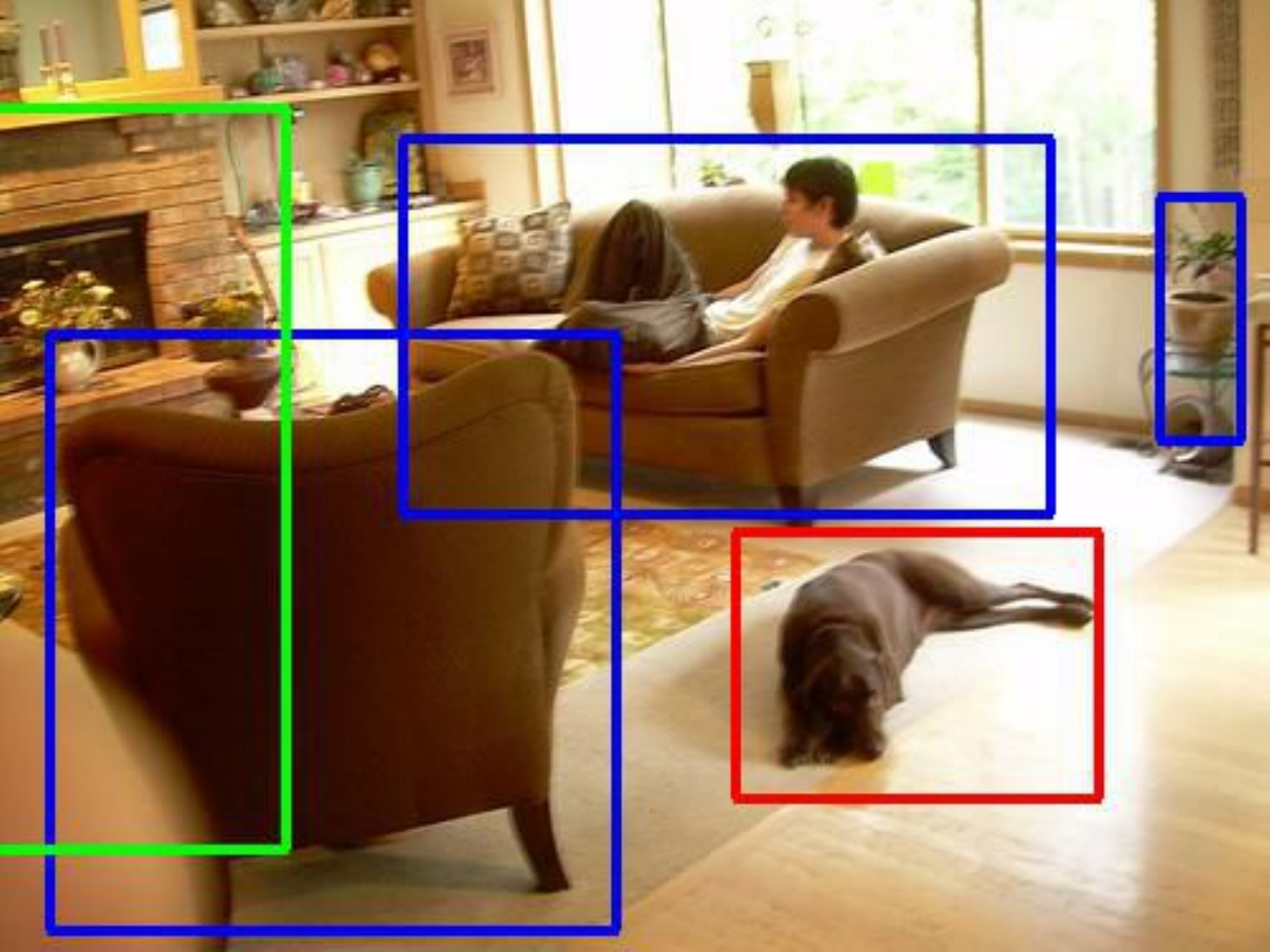}
\end{tabular}
\caption{Example of detected unseen class through attributes. Our model (Right) detects the chair, sofa and fireplace even though they are not in the annotated list of classes in PASCAL whereas YOLOv2 (left) only detects the seen class dog. \label{figure:image_motivation}}
\end{figure}

The fundamental issue is that, YOLOv2 tends to relegate {\it unseen} objects as background leading to missed detection of unseen objects. In our experiments we often observe significant recall drops when a detector~\cite{redmon2016yolo9000} is trained on {\it seen} classes but applied to {\it unseen} object classes with zero training samples. YOLOv2's low recall rate for unseen objects is altogether not surprising because YOLOv2 achieves high-performance, in part, by implicitly learning unique object visual features that can be used to match true objects at test-time through penalization of objectiveness confidence (to suppress false positives and maximize precision) during training. As a result, if an {\it unseen} class does not share the visual features learned on training data, these unseen objects have low objectiveness scores resulting in low recall. %

{\it Key Insight:} We posit that we must explicitly train detectors that account for semantic attribute information in the visual domain to ensure that semantically similar object attributes will be reflected in the learnt visual features. In this way, at test-time, we can hope to detect unseen objects that are semantically related to seen object classes. 

{\it Recognition vs. Object Detection:} We focus on the problem of proposing object bounding boxes in presence of seen and unseen classes, although, our method can be readily extended to recognition if semantic information for unseen objects were available. We focus on object detection out of necessity (limitations of available datasets) as well as practical reasons (for objects in the wild scenarios semantic information for unseen classes is not available).

To train ZS-YOLO, we learn an end-to-end detection network with a hierarchical architecture (Fig.~\ref{figure:arc}): In the first level, we train the network with a multi-task loss to perform (1) bounding box prediction in the visual domain; and (2) attribute prediction in the semantic domain; Next, the visual features of each bounding box proposal and its semantic attributes are further combined as a multi-modal input for the final layer to produce an objectiveness confidence score. This setup must be contrasted with existing detection frameworks which predict the confidence score based solely on the visual space. Extensive experiments are conducted to verify the ZS-YOLO performance on both traditional zero shot detection where only unseen objects exist and generalized zero shot detection (GZSD) where both seen and unseen objects appear in the images. Our experiments evidently shows the benefits of our multi-task training and multi-modal confidence prediction strategy: we improve the recall rate of the baseline model~\cite{redmon2016yolo9000} from $18.6\%$ to $48.2\%$ on {\it unseen} classes at 0.9 confidence, and {\it unseen} average precision from $56.4\%$ to $60.1\%$ on {\it PASCAL VOC} dataset for 10(seen)/10(unseen) split. Similar trends are also observed on different data splits as well as {\it MS COCO} dataset. We then perform extensive ablative analysis and quantify the utility of semantic information in for unseen object detection. We then identify cases where attribute information is particularly useful for unseen object detection. 

Our {\bf contributions} in this paper are: (1) Novel method for zero-shot detection problem that seamlessly integrates semantic attribute predictors with visual features during training. %
 (2) Dataset: We construct a new ZSD dataset with multiple {\it seen} and {\it unseen} classes splits based on existing {\it PASCAL VOC} and {\it MS COCO} dataset; New performance metrics are also introduced and discussed; (3) We develop a new ZSD detector, based on the visual network structure of YOLOv2~\cite{redmon2016yolo9000}. In contrast to state-of-art detectors, ZS-YOLO learns to predict semantic attribute as a side task during training, and produces object bounding boxes with both visual and semantic information. We observe significant improvements on both {\it PASCAL VOC} and {\it MS COCO} for {\it unseen} classes.
 
\section{Related Work}
\subsection{Object Detection and Proposal}
We utilize YOLOv2~\cite{redmon2016yolo9000} as our baseline model, since it is fast ($>50$ FPS detection speed), simple (single-shot detection with a fully convolutional network structure~\cite{long2015fully}), and effective (state-of-art performance).
As a single-shot detector, YOLOv2 directly infers from image cells and simultaneously produces a fixed set of bounding box proposals together with their associated confidence scores.
Compared to YOLOv2, while other contemporary deep detectors can be as effective, they are less efficient (usually $<30$ FPS). For instance, Faster-RCNN~\cite{ren2015faster} and R-FCN~\cite{dai2016r} use a Region Proposal Network (RPN) as a parallel branch to first generate object proposals, and use the pooled region features to further refine bounding box locations as well as their objectness scores, which is usually much slower compared to YOLOv2.
Nevertheless, in the context of large-scale detection problems, these methods including YOLOv2 are somewhat ineffective in detecting unseen object classes when trained with no corresponding training data. This can be attributed to seeking high supervised performance during training, a strategy that leads to maximizing detection precision with seen classes and encourages the network to suppress any image regions with divergent visual features as false positive proposals. This strategy in turn hampers unseen object detection since proposals of novel or unseen objects are often relegated as background.

On the other hand, in contrast to deep detectors, which are trained to suppress false positives, region proposal methods~\cite{zitnick2014edge,zhang2015bing++,uijlings2013selective,ren2015faster} can potentially discover objects that are unseen. However, region proposal methods suffer from significantly high false positive rates. While they are designed to propose hundreds of regions per object, this leads to poor precision and requires significant computational resources for post-processing to improve accuracy. For these reasons we leverage YOLOv2 rather than improve precision of object proposal methods. 
Specifically,  we extend YOLOv2 by leveraging the semantic attributes in the detection architecture. Nevertheless our method can be applied to other detectors, but we choose to demonstrate it on YOLOv2 due to superior performance.

\subsection{Zero Shot Recognition}
Zero Shot Learning (ZSL) seeks to recognize novel visual categories that are unannotated in training data \cite{lampert2014attribute,Xian_2016_CVPR,lei2015predicting,zhang2016zero}.
Traditional ZSL is constrained to closed set of unseen classes only. Recently it has been shown that traditional methods do not generalize well to the case where classification includes both seen and unseen classes at test time \cite{Chao2016,xian2017zero,fu2017recent}. While generalized ZSL (GZSL) relaxes this constraint that the test data only belongs to unseen class, it still does not deal with background class which significantly increases the problem complexity. Our method complements the existing ZSL work by reducing the open set problem to a closed set one by detecting only the foreground objects. After bounding boxes of unseen objects are extracted by our detector, any ZSL method can be applied to obtain object labels.

\subsection{Other Methods}
Recent work on weakly supervised localization have proposed methods to localize objects without bounding box annotations \cite{oquab2015object,zhou2014object,Bilen_2016_CVPR}. However, these methods still rely on image level object annotations and are not focused on detecting the unseen. In our case, there are no annotations for unseen classes in training data.
There are also methods that aim to discover objects without supervision \cite{cho2015unsupervised}. These methods usually rely on redundant parts and features in the training data to discover patterns and clusters of objects. Our goal is to transfer the detection knowledge of seen classes through semantic attributes to unseen ones. Fully unsupervised methods do not utilize the semantic transfer and may have hard time discovering clusters that are completely missing in the training data.

\subsection{Concurrent Works}
Two other groups have concurrently worked on ZSD~\cite{bansal2018zero, rahman2018zero} and appeared around the time a preliminary version of this paper was posted on arxiv. However, both approaches differ significantly from ours. Both works assume object proposals are predicted by a predefined proposal generator (Edge-Box~\cite{zitnick2014edge} for \cite{bansal2018zero} and Region Proposal Network(RPN) for \cite{rahman2018zero}), and focus on the subsequent ZSR problem, which is to map the visual features extracted from object proposals to a semantic embedding and perform classification. Nevertheless, this assumption of already available proposals on unseen objects is unrealistic in the GZSD problem. In fact, one key argument of this paper is that traditional detectors/proposal generators often suppress unseen objects as backgrounds, and thus cannot detect unseen objects initially. To mitigate this issue, this paper proposes a novel confidence prediction layer which takes a combination of visual features, semantic features, and spatial locations as input to jointly justify the existence of unseen objects. In addition, unlike \cite{bansal2018zero,rahman2018zero} that exploit a two-step detection setup, our GZSD detector is built on the one-step YOLO detector, which is fast and scalable to large data size. Last but not the least, this paper adopts a substantially more thorough evaluation setting with multiple splits of both ZSD and GZSD experiments, whereas both \cite{rahman2018zero} and \cite{rahman2018zero} are mainly evaluated on the ZSD test setting (only unseen classes) with only one split. We also noticed that \cite{rahman2018zero} only evaluated on the ILSVRC-2017~\cite{park2017ilsvrc} which is more constrained where most images contain only one object that needs to be detected. 

\section{Methodology}

\subsection{Problem Definition}
First, let us formally define the problem of detecting unseen object classes with zero training data, or Zero-Shot Detection (ZSD). Assuming a detection training set of $N$ ground truth bounding box labels is given as $\mathcal{D}_{tr} = \{ \mathbf{B}_i, c_i,  \mathbf{y}_i\}_{i = 1}^N$, where $\mathbf{B}_i = \{x_i, y_i, w_i, h_i\}$ specifying the location and size of each object bounding box, $c_i \in \mathbf{C}_{seen}$ denoting the class label of the $i$-th object, and $\mathbf{y}_i \in \mathbb{R}^h$ is its corresponding $h$-dimensional semantic representation vector, e.g. attributes. During testing time, we assume images contain objects with both seen classes $\in \mathbf{C}_{seen}$  and unseen classes $\in \mathbf{C}_{unseen}$ ($\mathbf{C}_{seen} \cup \mathbf{C}_{unseen} = \varnothing$). The task of ZSD is {\it to recognize every foreground objects against backgrounds and meanwhile predict their bounding boxes}.

{\it When developing ZS-YOLO, we deliberately do not require it to identify the test object class names for three reasons:}
(1) The failure of YOLOv2 on unseen classes is caused by missing detection rather than classification. That is, YOLOv2 tends to relegate unseen classes as backgrounds. Thus, the low recall rates on unseen classes becomes the main issue for ZS-YOLO to address;
(2) Once the missing detection problem is solved, the classification task of seen and unseen classes with the extracted bounding boxes has already been extensively explored by ZSR with saturated off-the-shelf solutions~\cite{zhang2015zero,zhang2016zero,lampert2014attribute,lei2015predicting}; 
(3) Different to ZSR where the set of unseen classes $\mathbf{C}_{unseen}$ are given during training, ZS-YOLO works in a more generalized setting where $\mathbf{C}_{unseen}$ in completely unknown and unconstrained. In other words, its task is to detect any foreground objects whose classes are not limited by any pre-defined set. 

\subsection{ZS-YOLO Network Architecture}
\begin{figure*}[t]
\centering
\includegraphics[width=\textwidth]{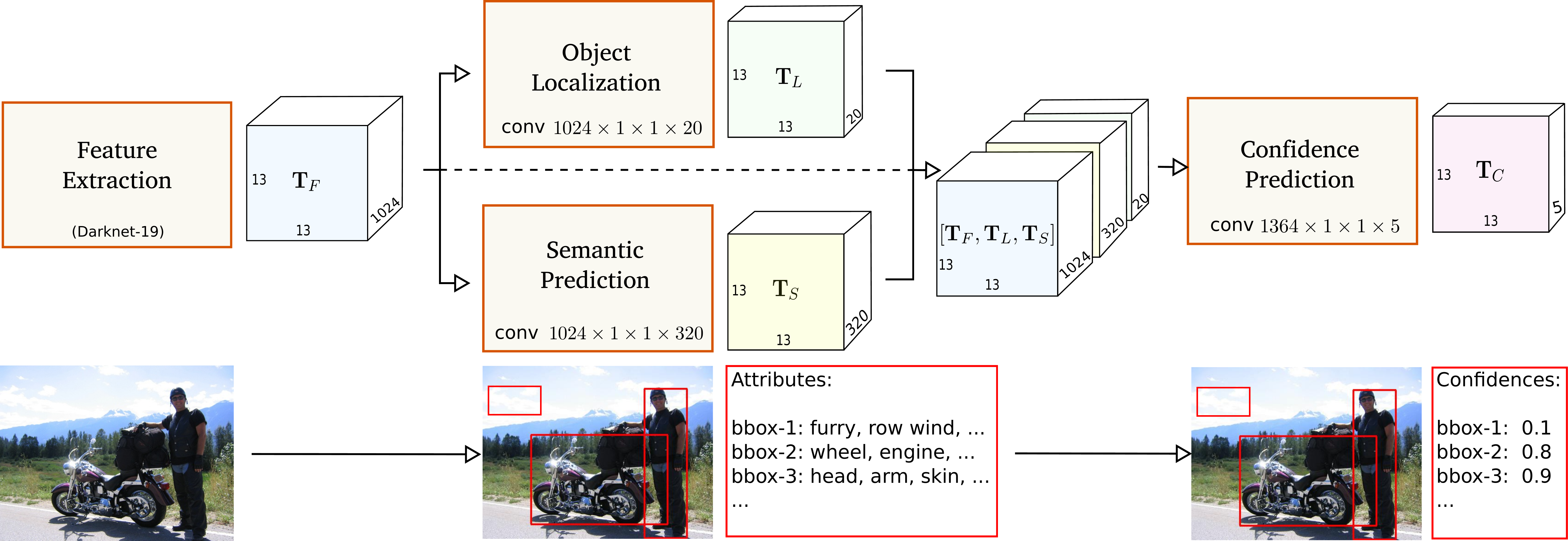}
\caption{The network architecture of our ZSD model. We choose $h = 64$ for our experiments on {\it PASCAL VOC} dataset, so the output dimension of the Semantic Prediction layer is $64 \times 5 = 320$. In the block ($c \times w \times h \times d$) represent the size of the convolution filter, where $c$ = input channel, $(w, h)$ = filter size, $d$ = output channel.}
\label{figure:arc}
\end{figure*}

The architecture of the proposed ZS-YOLO is shown in Fig.~\ref{figure:arc}. It consists of four modules: a feature extraction module, an object localization module, a semantic prediction module, and an objectiveness confidence prediction module. Same as YOLOv2~\cite{redmon2016yolo9000}, ZS-YOLO is fully convolutional~\cite{long2015fully}, i.e. constructed by only convolutional layers and pooling layers. 

\subsubsection{Feature Extraction}
We take the backbone architecture of YOLOv2~\cite{redmon2016yolo9000} (named as Darknet-19) as the CNN subnet due to its superiority on handling multi-scale objects with a specifically designed passthrough layer applied on fine-grained features. In practice, any other popular CNN model, e.g. ResNet~\cite{he2016deep}, Inception-V3~\cite{szegedy2016rethinking}, VGG16~\cite{simonyan2014very}, can be replaced here as the base network.
Same as \cite{redmon2016yolo9000}, we resize input images into $416 \times 416$, and our feature extractor outputs a feature tensor, denoted as $\mathbf{T}_F$, in shape of $13 \times 13 \times 1024$ for each input image.
After feature extraction, our network is then divided into two branches, which perform object localization and semantic attribute prediction respectively. 

\subsubsection{Object Localization}
Similar to \cite{redmon2016you}, we divide each image into $13 \times 13$ cells, each represented by a $1024$-dim vector in $\mathbf{T}_F$. Each cell is assigned with 5 anchor boxes with pre-defined aspect ratios, and needs to predict 4 coordinate offsets for each anchor box. This is implemented by a fully convolutional layer with $4 \times 5 = 20$ filters of $1 \times 1$ kernel size. The output tensor, denoted as $\mathbf{T}_L$, is thus in shape of $13 \times 13 \times 20$. For locating a predicted bounding box, assuming the cell location is denoted by $(c_x, c_y)$, the width and height of an anchor box is $(p_w, p_h)$, and the predicted offsets are $(o_x, o_y, o_w, o_h)$, $\sigma(\cdot)$ the sigmoid function, then the location of this prediction is calculated by:
\begin{align}
\hat{x} &= \sigma(o_x) + c_x, & \hat{y} &= \sigma(o_y) + c_y, \nonumber \\
\hat{w} &= p_w e^{o_w}, & \hat{h} &= p_h e^{o_h}
\end{align}

\subsubsection{Semantic Prediction}
In addition to the bounding box locations, we assign each cell a new task to predict the corresponding semantic representation $\hat{\mathbf{y}} \in \mathbb{R}^h$ (e.g. attributes) for every bounding boxes proposals. Specifically, since each cell proposes 5 bounding boxes, the output tensor $\mathbf{T}_S$ is thus in shape of $13 \times 13 \times 5h$, which again can be implemented by a convolutional layer with $5h$  filters of size $1 \times 1$. $h$ is determined by the semantic annotations and may vary across datasets. For example, on {\it PASCAL VOC} we use the $64$-dim attributes released by \cite{farhadi2009describing} as the semantic representations, so $h$ is set to be 64. Training the semantic prediction side-task brings two advantages: (1) For each testing bounding box proposal we can obtain its semantic representation; (2) As ZS-YOLO is trained end-to-end, the loss of this layer will back propagate so that the learned visual representations will also be influenced by similarities in the semantic domain. Both the learned semantic and visual representations will be then incorporated for the further objectiveness confidence prediction. 

\noindent \subsubsection{Detection of Overlapping Objects.} Observe that our method allows for detection of overlapping objects either from the same or of different classes. This is because each bounding box outputs its own attribute prediction. Therefore, even if the objects are overlapping, such as the motorbike and person in Fig. \ref{figure:arc}, the predicted attributes correspond to different bounding boxes and the outputs are different, enabling recognition. Our semantic loss (discussed later) can further enforce diversity among predictions. In addition, even when bounding boxes are in the same location, as a consequence of YOLOv2, they arise from different cells, and so the attributes will be learned from different visual features resulting in different outputs.

\noindent
\subsubsection{End-to-End vs. Two-Step  Recognition.\label{sec:endtoend}} Our model could be viewed as an embedding based end-to-end zero-shot localization and recognition method, when semantic attributes for all target classes are available. This follows from the fact that our model produces two outputs at test-time, a bounding box with visual features, and a semantic output. The semantic output is not used whenever target class semantic attributes are unavailable, such as in-the-wild recognition problems. Nevertheless, observe that the semantic output for test examples is always available, and incorporates both test visual features as well as semantic attributes for seen classes. Consequently, the semantic output can be directly compared against ground-truth semantic attributes such as nearest-neighbor or cosine similarity to output a label. In an embedding method, the visual features are mapped to semantic space and these attributes are compared against ground-truth. Viewed in this way, ours is an embedding based method~\cite{xian2017zero}, a standard approach for generalized zero-shot learning. Compared to the cascaded or two-step ZSR methods, the computation of the NN classifier is negligible, and the performance is similar to a two-step approach, as we show in the experiments. Moreover, other loss functions widely used in zero shot recognition methods (e.g. max-margin losses) may also be leveraged for the semantic prediction for our setting.

\subsubsection{Confidence Prediction}
The final component of our network is to predict a confidence score, $\hat{p}$, associated with each bounding box proposal. Existing detectors, e.g. SSD~\cite{liu2016ssd} and YOLOv2~\cite{redmon2016you,redmon2016yolo9000}, predict the confidence score directly from the CNN feature map $\mathbf{T}_F$. However, such a strategy produces low confidences for {\it unseen} objects visually different to the {\it seen} training data, and thus suffering from low recalls. To address this problem, ZS-YOLO predicts the confidence utilizing information from both visual and semantic domain. Additionally, the bounding box coordinates can also be explored as useful sources for predicting the objectiveness confidence, since foreground objects are often located on certain locations (e.g. center rather than corners) of an image.

More specifically, we concatenate all output tensors from previous modules ($\mathbf{T}_F$, $\mathbf{T}_L$, $\mathbf{T}_S$) as a combined multi-modal input tensor with shape $13 \times 13 \times (1024 + 20 + 5h)$, and forward it into a convolutional layer with 5 single dimensional filters, generating an $13 \times 13 \times 5$ output tensor $\mathbf{T}_C$, each bit representing the predicted confidence score $\hat{p}$ for the corresponding one of the $13 \times 13 \times 5 = 845$ bounding boxes on an image.

\subsubsection{Choice of YOLOv2 Detection Architecture.}
In our implementation we choose YOLOv2 as the backbone network for three reasons: (a) As a single stage detector, YOLOv2 has a very similar structure with SSD\cite{liu2016ssd}. The only major difference is that YOLOv2 uses Darknet as the feature extractor while SSD uses VGG. As reported in \cite{redmon2016yolo9000}, YOLOv2 reaches higher accuracy on VOC2007 test set while the speed is also faster than SSD512. It is thus not difficult to apply the same idea on SSD. (b) Another popular single stage detector, RetinaNet\cite{lin2017focal}, mainly leverages focal loss. Our experiments with penalizing focal loss did not demonstrate noticeable benefits on performance, as we will discuss later in experiments. (c) Many two-stage detectors, like Faster-RCNN\cite{ren2015faster}, are based on the so called Region Proposal Network (RPN), which generates bounding box proposals in the first stage. This two-stage setup makes the detector very slow compared to single stage network\cite{redmon2016yolo9000}. The RPN outputs the bounding boxes and the confidence scores without doing any classification. Thus RPN could be seamlessly integrated into ZS-YOLO by substituting the detector module in ZS-YOLO with an RPN module. Nevertheless, while these offer a number of options, our focus here was to understand and quantify the benefits of semantic attributes for zero-shot detection, and less on comparison of different detectors.

\subsection{Zero-Shot Detection Losses}
With the network architecture defined, we minimize an objective function with a multi-task loss specifically designed for ZSD. Our overall objective loss function is a weighted sum of an object localization loss, a semantic loss and a a confidence loss.

\subsubsection{Object Localization Loss}
Our network predicts multiple bounding boxes per grid cell, and we want only one bounding box predictor to be responsible for each object to avoid redundancy. We thus define an object indicator $\mathbb{I}_k^{obj}$ for every bounding box prediction. Assuming the $k$-th bounding box is preducted by cell $j$, $\mathbb{I}_k^{obj}$ is set to $1$ if and only if: (1) The center of a ground truth object annotation falls into cell $j$; and (2) Prediction $k$ has the highest IOU with the ground truth box among all the 5 predictions made by cell $j$. Otherwise $\mathbb{I}_k^{obj}$ is set to $0$. The localization loss for the $k$-th bounding box prediction is then defined as a sum-squared error between the predicted coordinates and its ground truth labels:

\begin{align}
L_{loc} &= \sum_{k = 1}^{845} \mathbb{I}_{k}^{obj} \big[ (\hat{x}_{k} - x_{k})^2 + (\hat{y}_{k} - y_{k})^2  \nonumber \\
&+ ( \sqrt{\hat{w}_{k}} - \sqrt{w_{k}})^2 + (\sqrt{\hat{h}_{k}} - \sqrt{h_{k}})^2 \big]
\end{align}

\subsubsection{Semantic Loss}
The semantic loss is designed so that our network could learn a semantic vector representation for each {\it seen} class attributes which could be generalized to {\it unseen} classes during testing time. 

 Given the ground truth semantic vectors ($\{\mathbf{y}_c\}, c \in \mathbf{C}_{seen}$) for each {\it seen} class, i.e. {\it prototypes}, our objective is to maximize the similarity of each semantic predictions on foreground objects to ground truth labels. Furthermore, to increase the network's discriminative capability, we also want the semantic predictions on background boxes to have low similarity with {\it seen} class prototypes. Similarly to our approach above, we define a background indicator $\mathbb{I}_k^{noobj}$ for every bounding box prediction. Assuming the $k$-th bounding box is predicted by cell $j$, $\mathbb{I}_k^{noobj}$ is set to $1$ if and only if cell $j$ does not overlap with any ground truth boxes. Otherwise $\mathbb{I}_k^{noobj}$ is set to $0$. During training, the number of background bounding boxes is much larger than foreground. To balance it, we impose the weights $\lambda_{obj}$ and $\lambda_{noobj}$ and set them to 5 and 1 in our experiments, respectively. Assuming $S(\mathbf{a}, \mathbf{b})$ denotes the cosine similarity between two vectors, i.e. $S(\mathbf{a}, \mathbf{b}) = \frac{\mathbf{a} \cdot \mathbf{b}}{\|\mathbf{a}\|_2 \|\mathbf{b}\|_2}$, 
our semantic loss is defined as:

\begin{align}
L_{attr} &= \sum_{k = 1}^{845} \big[ \lambda_{obj} \mathbb{I}_{k}^{obj} (S(\hat{\mathbf{y}_k}, \mathbf{y}_k) - 1)^2 \nonumber \\
&+ \lambda_{noobj} \mathbb{I}_k^{noobj} (\max_{c \in \mathbf{C}_{seen}} S(\hat{\mathbf{y}_k}, \mathbf{y}_c) - 0)^2\big]
\end{align}

\subsubsection{Confidence Loss}
Finally, the confidence loss is imposed so that our network will predict high objectiveness confidence scores on foreground bounding box proposals and low scores for bounding boxes containing only image backgrounds. Formally our confidence loss is defined as:

\begin{align}
L_{conf} &= \sum_{k = 1}^{845} \big[ \lambda_{obj} \mathbb{I}_{k}^{obj} (\hat{p}_i - 1)^2 
+ \lambda_{noobj} \mathbb{I}_k^{noobj} (\hat{p}_i- 0)^2\big] \label{eqn:conf_loss}
\end{align}

Note that we put a minus 0 explicitly in Eq.(3-4) to emphasize the training objective that when the predicted bounding box contains only backgrounds, the max semantic similarity score and the confidence score should be close to zero.

Finally, for each training image, its total objective loss is calculated by a weighted sum of all the above three losses with weights $\lambda_{loc}, \lambda_{attr}, \lambda_{conf}$. In our experiments, we set all of them to be 1. Our network is trained by evaluating the empirical loss over the entire training set via stochastic gradient decent.

\subsection{Training Details}
We use the first 16 convolutional layers from Darknet-19~\cite{redmon2016yolo9000} pretrained on {\it ImageNet 1K-class} dataset~\cite{deng2009imagenet} and add 3 randomly initialized convolutional layers as our feature extractor. 

{\it Validity of Pretraining Weights for Detection.} Note that the pretrained weights correspond to the Darknet weights on ImageNet classification task, and not the detection challenge task. Thus no localization supervision is provided in pretraining, which is the focus of this paper. Also, when training the detector, no labels or bounding boxes for unseen objects are accessible. Therefore, observe that the visual features for unseen classes are viewed as background unless they are learned from seen classes. Consequently, the representation embedded in the pretrained weights will tend to be suppressed due to lack of labels. In addition, in the experiments, both YOLOv2 and ZS-YOLO are initialized with pretrained Darknet weights for fair comparison. We also point out that our focus here is to understand benefits of semantic attributes in the best situation, where other components, modules and weights are suitably well-chosen. Indeed, the general problem of jointly optimizing the entire system is important but out of scope of this paper. Based on what we observe the issue of the ImageNet pretrained weights does not appear to be a dominant aspect of the large performance gap between seen and generalized zero-shot detection.

The activation functions of the semantic predictor and confidence predictor are linear and all other layers use a leaky rectified linear activation function $\Phi(x)$:
\begin{equation}
\Phi(x) =
\begin{cases}
x & \text{if } x > 0\\
0.1x & \text{otherwise}
\end{cases}
\end{equation}

The network is trained end-to-end for 420 epochs on the training splits from {\it PASCAL VOC} 2007 and 2012 and {\it MS COCO}. We use a SGD optimizer with the batch size 64, momentum 0.9 and a decay of 0.0005. As for the learning rate, in the first 5 epochs, we set it to 0.0001 because the model often diverges if it starts with a high learning rate. Then we increase it to 0.001 and train the model for 195 epochs. Afterwards we decrease it to 0.0001 for 110 epochs, and train it with 0.00001 for the final 110 epochs.

\section{Experiments}
For our evaluation metrics, we measure the detection performance by Average Precision (AP), which is the average precision of all classes in the dataset. For our evaluation metrics, we measure the Pascal VOC~\cite{everingham2010pascal} 0.5-IOU 11-point average precision (AP). Specifically, the ground truth overlapped by a predicted bounding box with IOU over 0.5 is counted as True-Positive (TP); the number of ground truth objects is $\mathrm{GT}$; and $\mathrm{Pred}$ denote the number of predicted bounding boxes. For a specific confidence threshold, the precision (Prec) and recall (Rec) is computed via:
\begin{align}
    \mathrm{Prec} &= \mathrm{TP} / \mathrm{Pred}\\
    \mathrm{Rec} &= \mathrm{TP} / \mathrm{GT}
\end{align}
The average precision is then defined as the mean precision at a set of eleven equally spaced recall levels $[0, 0.1, 0.2, ...]$:
\begin{equation}
\mathrm{AP} = \frac{1}{11} \sum_{r\in[0, 0.1, ...]} \mathrm{Prec}(\mathrm{Rec}=r)
\end{equation}
Unlike standard mAP used in Pascal VOC, we measure AP over all classes since we do not classify the objects. There is no way to get a class-specific precision and so class-level AP and mAP are not computed. The AP reflects the overall detection performance on all possible objects. Nevertheless, as we have already mentioned in Sec.~\ref{sec:endtoend}, we could conceivably use our system as an embedded based recognition method, whenever semantic attributes for ground-truth are provided. With this in mind we also tabulate recognition results for the sake of completeness.

Nevertheless, we specifically care about AP of unseen classes since our goal is to leverage YOLOv2 to detect more unseen objects. We also report the average F-score as an auxiliary measurement since it reflects the average performance of the detector at different confidence thresholds The average F-score is computed by averaging the F-scores over a set of 101 equally spaced confidence thresholds $[0, 0.01, 0.02, ...]$, at each confidence threshold, the F-score is defined as:
\begin{equation}
    \text{F-score} = \frac{\mathrm{Prec}\cdot\mathrm{Rec}}{\mathrm{Prec} + \mathrm{Rec}}
\end{equation}
The average F-score can reflect the robustness of overall performance for the detector.

\subsection{Datasets and Settings}
\label{sec:setting}
To evaluate the zero-shot detection performance of the proposed ZS-YOLO model, we chose {\it PASCAL VOC}~\cite{everingham2010pascal} 2007 and 2012 datasets and {\it Microsoft COCO} \cite{lin2014microsoft} object detection dataset due to their popularity among object detection literature. 

\begin{table}[t]
\centering
\setlength{\tabcolsep}{0.3cm}
\begin{tabular}{lccccc}
\hline
Data & VOC2007 & VOC2012 & COCO & Class\\
\hline
Train & train/val & train/val & train & seen\\
Test-Seen & test & - & val & seen\\
Test-Unseen & train/val+test & train/val & train/val & unseen\\
Test-Mix & train/val+test & train/val & train/val & both\\
\hline \\
\end{tabular}
\caption{Our data splits on PASCAL VOC and MS COCO dataset.}
\label{data_component}
\end{table}

\subsubsection{PASCAL VOC Dataset} This dataset contains 20 object classes in total, and each class is labeled with $64$-dimensional semantic attributes published by ~\cite{farhadi2009describing}. The binary instance level attributes are provided in \cite{farhadi2009describing} and we compute the class-level attributes by averaging over all instances in the class. There is no standard seen/unseen split on {\it PASCAL VOC} for object detection, we thus built our own splits. Since PASCAL VOC is a relatively small dataset we utilize PASCAL primarily for several ablative studies. In our first experiment our goal is to quantify the impact of increasing unseen classes. We discuss other ablative studies in Sec.~\ref{sec:analysis}. For this study, we held out different numbers of unseen classes and utilized the rest for training. These unseen classes were selected based on their diversity in class types (e.g. avoiding similar classes such as {\it dog} and {\it cat} from being both included as unseen). We thus ended up with three different seen/unseen class balances: respectively 15/5, 10/10, and 5/15. 
During model training for each split, we collect all the images which only contain seen object classes in the {\it train/val} partition of {\it PASCAL VOC} 2007 and 2012 dataset as training data.
For testing, we use three different data configurations, named as {\bf Test-Seen}, {\bf Test-Unseen} and {\bf Test-Mix}.
Test-Seen data is constructed by the images from VOC2007 test partition which only contain seen classes; 
Test-Unseen data contains all the images having only unseen class objects from both {\it train/val/test} partitions of VOC2007 and {\it train/val} partition of VOC2012; 
All the other images, which contain both seen and unseen objects, go into our Test-Mix data. Test-Unseen corresponds to the standard zero shot detection task, where only unseen classes are in the data. The ability of discovering unseen objects can be evaluated on Test-Unseen. Test-Mix is a more sophisticated situation where both seen and unseen objects appear in the image. The detector needs to identify unseen objects (assign high confidence score) even in the presence of seen objects that the detector has evidently been optimized to detect during training. The high score of seen objects may suppress the prediction on unseen objects resulting in poor unseen scores. Finally, Test-Seen is a conventional detection dataset and quantifies the ability to ensure good detection for seen (supervised) classes. We list the components of our dataset in TABLE~\ref{data_component} in more detail.

{\it Class-level Attributes vs. Object-level Attributes}: There are three reasons for us to adopt the averaged class-level attribute representation, instead of the original object-level attributes by ~\cite{farhadi2009describing}. (1) During the training stage of GZSD/R models, while the class-level averaging might result in reducing the distinctiveness for each object, the annotation noises and variations also decrease after averaging. As observed in our experiments, adopting the class-level attribute representation as ground-truth label makes the training procedure more stable and results in better GZSD performance (similar observations also made by previous ZSR works, e.g.\cite{akata2016label, zhang2015zero, akata2015evaluation}). This might be easy to understand since the task is to achieve a good GZSD/R performance globally, rather than to distinguish each instance within a specific class. (2) During the testing stage of GZSD/R models, only the class-level semantic information (attributes) of the unseen classes is provided, and the object-level unseen attributes are impossible to acquire beforehand. (3) Finally, the aPY dataset~\cite{farhadi2009describing} only labels a fraction of PASCAL VOC objects and there are still a large amount of objects in PASCAL VOC without any object-level labels.

\subsubsection{MS COCO Dataset} {\it MS COCO} dataset has 80 classes which include all 20 classes in {\it PASCAL VOC}. 
While  an  attribute  dataset for MS COCO has been published in \cite{patterson2016coco}, we could not utilize it for  two reasons:  (1) The dataset only labels 29 out of the 80 object classes.  The number is too limited to conduct extensive ZSD experiments on COCO. (2) Many attributes provided in \cite{patterson2016coco} are not visually meaningful, e.g., professional,useful, friendly, functional, and thus not suitable for visual detection/recognition.
An alternative to attributes, word2vec~\cite{mikolov2013distributed} (w2v), is noisy and performs poorly. To extend our model to train on more classes, we propose to learn transformation $\mathbf{P}$, that maps w2v features (300 dimensions) onto a lower-dimensional w2vR space (25 dimensions, i.e. $h=25$), which is constrained to mirror VOC attribute similarity on the 20 common classes between {\it MS COCO} and PASCAL VOC. Specifically, for class $i$ attribute $\mathbf{y}_i$, and w2v vector $\mathbf{W}_i$, we seek $\mathbf{P}$ such that $$\langle \mathbf{y}_i, \mathbf{y}_j \rangle \approx \langle \mathbf{PW}_i, \mathbf{PW}_j \rangle.$$ This problem can be solved by using any one of the rank-approximate methods. More detailed evaluation can be found in TABLE~\ref{table:diff_semantic}, Section \ref{sec:analysis}.

{\it MS COCO} has more classes, while PASCAL VOC has few classes and this offers several options for conducting ablative studies. First, with PASCAL VOC we can tabulate effect of different ratios of seen/unseen classes. For the largest ratio, we hope to see less difference between ZS-YOLO and YOLOv2 (for test-mix or test-seen) because visual features are significantly stronger than semantic information. In contrast for very small ratio there are more unseen classes and very few seen classes and so we hope to see generally poor performance. Second, for MS COCO, since there are a large number of classes we can conduct a different type of experiment, namely, how unseen performance varies as we see more seen classes. For this reason, we tabulate performance for a fixed set of unseen classes as a function of more seen classes available for training. Third, MS COCO also allows for us to study the difference between YOLOv2 and ZS-YOLO at a larger scale in the presence of significantly more training data.

Based on this motivation, we first manually selected 20 classes that were sufficiently diverse for unseen categories. We then chose $N$ training categories that were semantically most similar in word2vec representation to the unseen categories. 
The number $N$ was increased in increments of 20. Similar to {\it PASCAL VOC}, we collected all images which only contain seen object classes in the train partition of {\it MS COCO} dataset as training data. Accordingly, Test-Seen is constructed by images only containing seen objects in validation partition of {\it MS COCO}; Test-Unseen contains all images only containing unseen objects in train/val partitions of {\it MS COCO} and all the images containing both seen and unseen objects in {\it MS COCO} train/val dataset, go into Test-Mix data. The components of the dataset are detailed in TABLE~\ref{data_component}.

\begin{figure*}[h]
\includegraphics[width=0.24\linewidth]{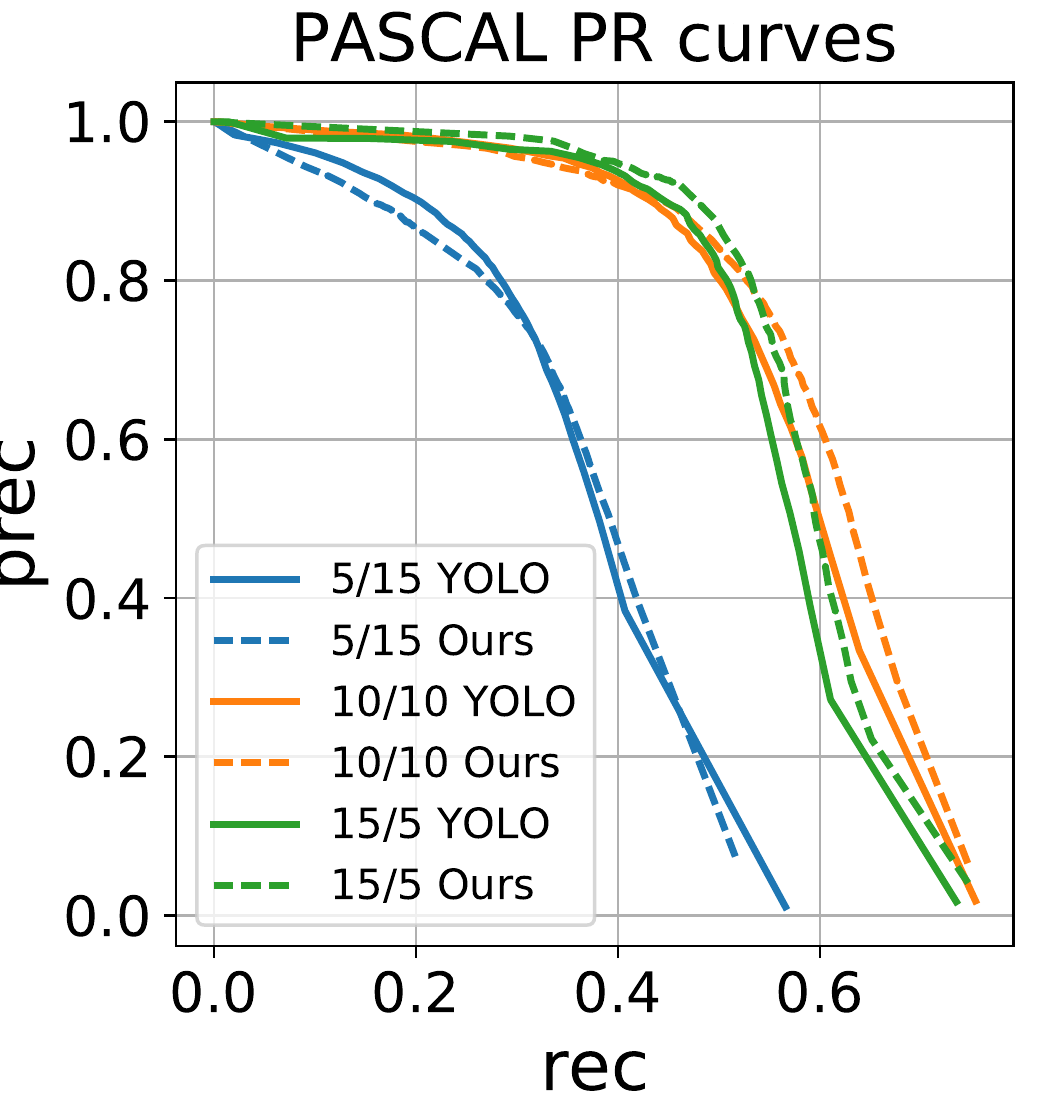}
\includegraphics[width=0.24\linewidth]{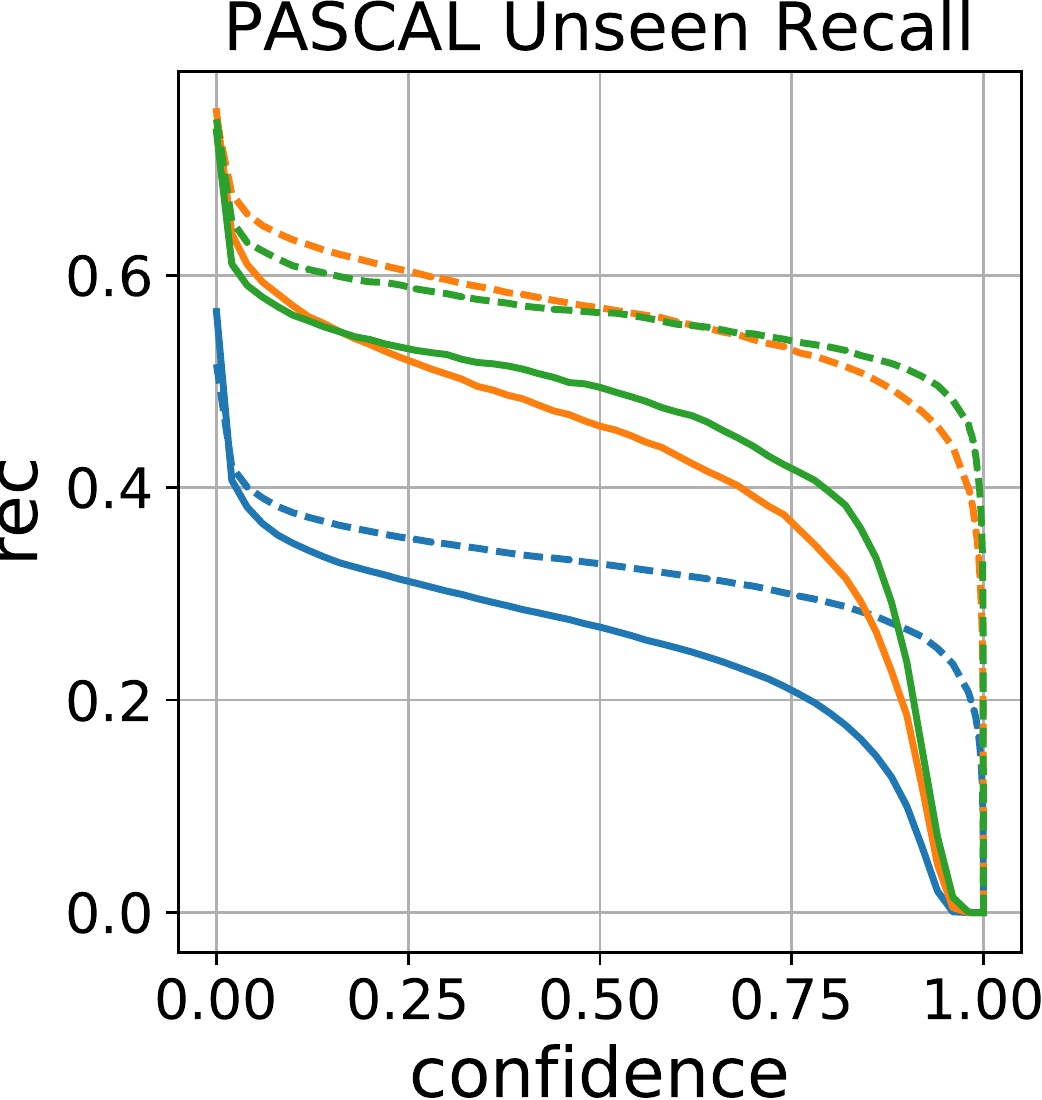}
\includegraphics[width=0.24\linewidth]{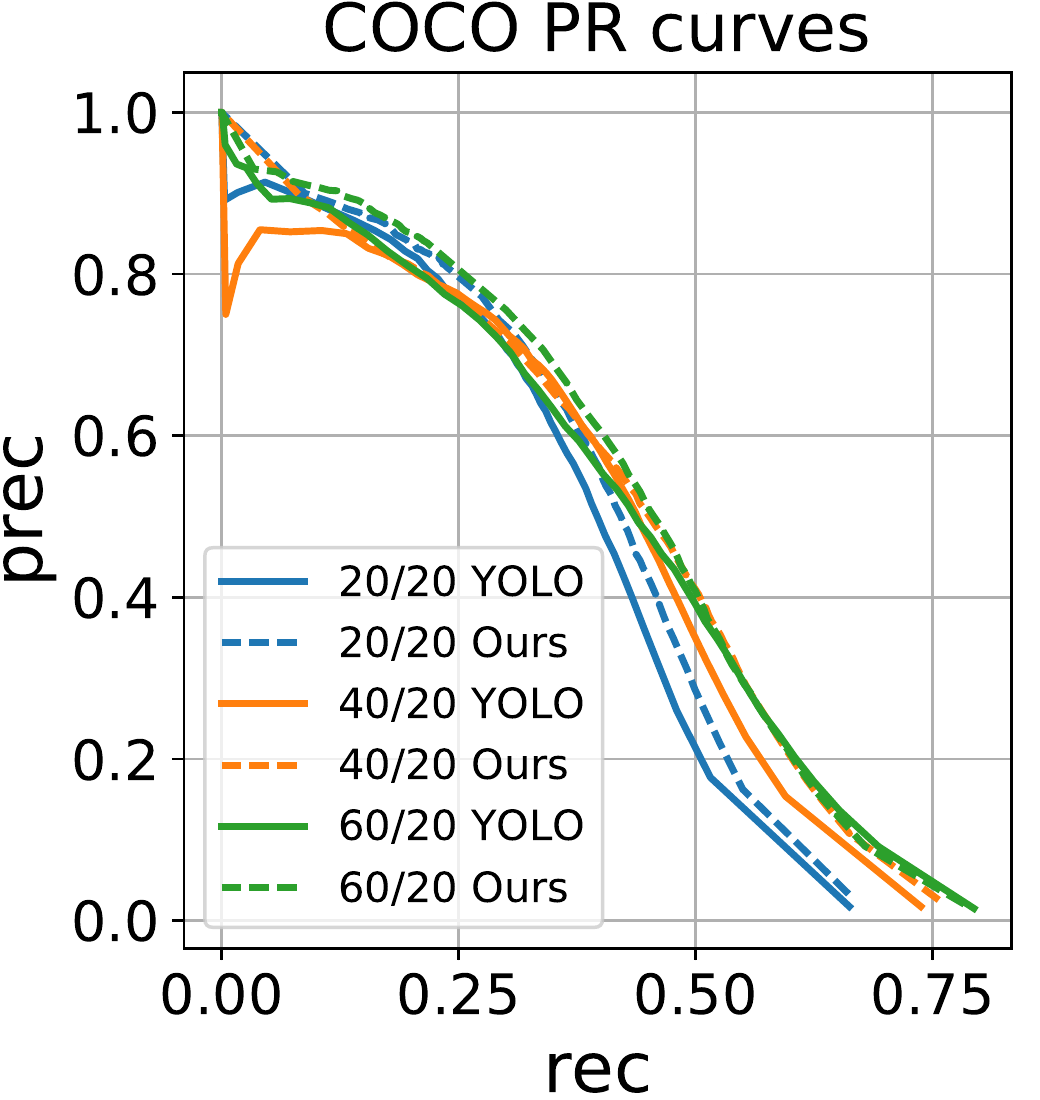}
\includegraphics[width=0.24\linewidth]{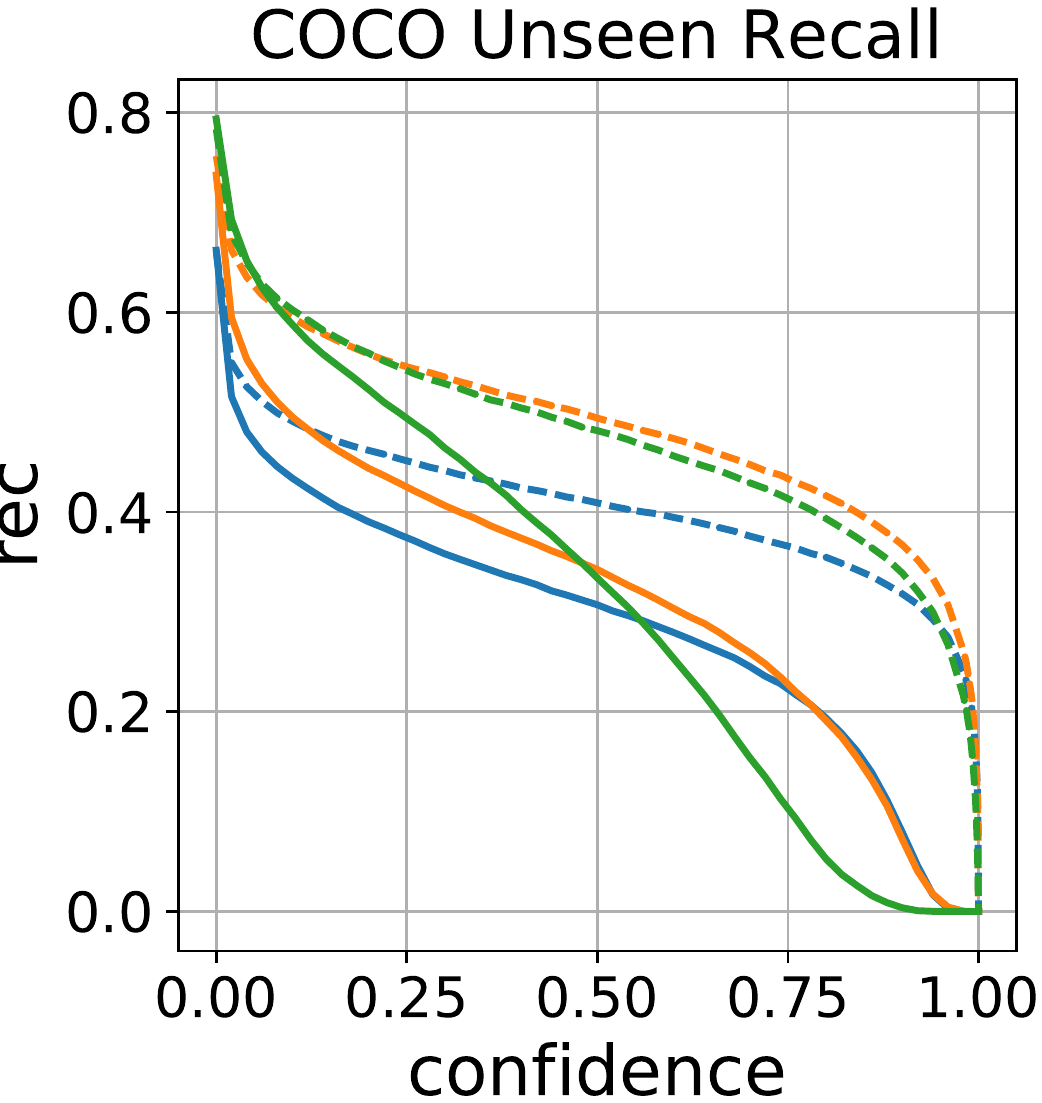}
\caption{Precision recall and recall curves of compared models on 15/5, 10/10 and 5/15 splits for unseen classes. Our model's recall exceeds YOLOv2 especially when the confidence is high.\label{fig:prec_recall}}
\end{figure*}

\begin{table}[t]
\centering
\renewcommand{\arraystretch}{1.1}
\setlength{\tabcolsep}{0.1cm}
\begin{tabular}{l | l l c c c c c c}
\hline
 & \multirow{2}{*}{Split}  & \multirow{2}{*}{Model} & \multicolumn{2}{c}{Test-Unseen} & \multicolumn{2}{c}{Test-Seen} & \multicolumn{2}{c}{Test-Mix}\\
 & & & AP(\%) & F-score & AP(\%) & F-Score & AP(\%) & F-score\\
\hline
\hline
\multirow{6}{*}{\rotatebox{90}{PASCAL VOC}} 
& \multirow{2}{*}{5/15}  & YOLOv2 & 36.6 & 22.1 & \bf{85.6} & 48.9 & 30.0 & 17.9\\
&        & ZS-YOLO & \bf{37.3} & \bf{37.3} & 85.0 & \bf{81.1} & \bf{30.9} & \bf{29.1} \\ \cline{2-9}
& \multirow{2}{*}{10/10}  & YOLOv2 & 56.4 & 24.3 & \bf{71.6} & 30.7  & \bf{54.3} & 23.6\\
&        & ZS-YOLO & \bf{60.1} & \bf{43.7} & 71.0 & \bf{52.3} & 53.9 & \bf{41.2}\\
\cline{2-9}
& \multirow{2}{*}{15/5}  & YOLOv2 & 55.3 & 33.0 & \bf{75.3} & 40.4 & 53.6 & 29.4\\
&        & ZS-YOLO & \bf{57.3} & \bf{56.9} & 73.9 & \bf{65.6} & \bf{53.8} & \bf{47.9}\\
\hline 
\hline
\multirow{6}{*}{\rotatebox{90}{MS COCO}} & \multirow{2}{*}{20/20}  & YOLOv2 & 37.3 & 22.0 & 34.5 & 25.9  & 12.3 & 12.4 \\
&      &  ZS-YOLO & \bf{40.6} & \bf{38.2} & \bf{41.2} & \bf{41.8} & \bf{20.2} & \bf{19.9} \\
\cline{2-9}
& \multirow{2}{*}{40/20} &  YOLOv2 & 40.8 & 22.8 & \bf{48.7} & 27.1 & 24.6 & 18.5 \\
&      & ZS-YOLO & \bf{42.7} & \bf{37.6} & 44.0 & \bf{42.0} & \bf{30.0} & \bf{28.7} \\
\cline{2-9}
& \multirow{2}{*}{60/20} & YOLOv2 & 34.9 & 23.6 & \bf{44.8} & 24.3 & \bf{37.6} & 19.8 \\
&       & ZS-YOLO & \bf{43.8} & \bf{35.3} & 40.6 & \bf{36.5} & 33.6 & \bf{29.3} \\
\hline
\multicolumn{3}{c}{}\\
\end{tabular}
\caption{Zero-Shot detection results on different datasets and Seen/Unseen splits. We perform significantly better than YOLOv2 for unseen objects while maintaining seen object performance.}
\label{table:diff_split}
\end{table}

\subsection{Zero-Shot Detection Evaluation}
\label{sec:splits}
We first evaluate the proposed ZS-YOLO model by comparisons on all seen/unseen splits as well as Test-Seen/Unseen/Mix configurations. As a baseline comparison, we also train a YOLOv2 as a standard fully-supervised detector (i.e. with both class and bounding box labels) using the same training splits. During test time, since we measure the average precision without classification, the classifier module in YOLOv2 is detached and the bounding box predictions are made based on confidence score.Our comparative results are shown in TABLE~\ref{table:diff_split} as well as Fig.~\ref{fig:prec_recall}. 

TABLE~\ref{table:diff_split} evidently shows the advantage of the proposed ZS-YOLO model, especially on detecting unseen object classes. Compared with YOLOv2, our model has a higher AP on Test-Unseen data on all the three different seen/unseen splits, e.g. we improve from YOLOv2's $56.4\%$ to $60.1\%$ on {\it PASCAL VOC} (10/10 split), and similarly, from $34.9\%$ to $43.8\%$ on {\it MS COCO} (60/20 split).
Our hypothesis is that the main reason for this performance gain in unseen AP is that ZS-YOLO predicts objectiveness confidence score based on both visual as well as semantic information, and thus effectively avoids suppressing unseen objects with novel visual features but closely-related semantic meanings. Observe that YOLOv2's performance is uneven as we increase seen class categories. For instance, on {\it MS COCO} dataset, YOLOv2 achieves the best performance on Test-Unseen ($40.8\%$) when trained on 40 seen classes, but worse when trained with 20 and 60 classes ($37.3\%$ and $34.9\%$ respectively). We attribute this to the fact to an under and over utilization of seen classes. Namely, in presence of few seen categories, without semantic knowledge, it is hard to generalize to unseen classes. On the other hand, in the presence of large seen training classes, YOLOv2 learns to reject unseen classes well and tends to classify unseen class as background. On the other hand, since ZS-YOLO exploits semantic feature to  overcome this issue, we observe a continuous improvements of ZS-YOLO on Test-Unseen as the number of training seen classes increases (from $40.6\%$ to $43.8\%$ on {\it MS COCO}). Additionally, we observe on Test-Seen partition that both models suffer from a performance drop when seen classes increase from 40 to 60. This is because many difficult classes are added (e.g. stop sign, remote, etc) which impede both models and lowers the average precision. 

From Fig.~\ref{fig:prec_recall}, observe that ZS-YOLO's recall on unseen objects is substantially larger than  YOLOv2 on both datasets thanks to its semantic attributes based detection framework. We illustrate this with the {\it PASCAL VOC} 5/15 split. Fig.~\ref{fig:prec_recall} shows that at a confidence threshold of $0.8$, ZS-YOLO can still detect about $30\%$ of the total unseen objects, whilst original YOLOv2's recall rate is only $20\%$. 
While ZS-YOLO achieves higher AP on unseen data with significantly improved recall rates, with degradation in precision on seen data. For instance, on {\it PASCAL VOC} ZS-YOLO loses $0.6\%$ (10/10), $1.4\%$ (15/5) and $0.6\%$ (5/15) AP on Test-Seen compared to YOLOv2. 
We posit that this is fundamental. Indeed, in order to detect more unseen data, we must inevitably accept more background bounding boxes to improve unseen object detection.  
However, if both models are evaluated with the Test-Mix data with both seen and unseen classes, we found that in general ZS-YOLO achieves better performance compared to YOLOv2.

We can observe from Fig.~\ref{fig:prec_recall} that except for VOC 5/15 split, ZS-YOLO has a higher precision than YOLOv2 in any level of recall. The 5/15 split is a special case where the number of seen attributes are somewhat insufficient to generalize to unseen objects.
For example, on VOC 10/10 split, at the recall level 60\%, YOLOv2 has precision 50\% while ZS-YOLO is 61\%. Also, ZS-YOLO has higher recall than YOLO on any level of confidence score. Although YOLOv2 achieves the same recall as ZS-YOLO at a smaller confidence threshold, it suffers from loss on precision. Consequently, we do benefit from semantic predictions, specifically, when we have sufficient number of training classes and, as noted later, when the semantic unseen and seen attributes are related Table~\ref{table:attribute_corr}. Noticeably, ZS-YOLO is robust, with a high recall in a wide range of confidence, meaning its performance does not degrade for a wide range of confidence scores. This can also be justified by the average F-score in Table.~\ref{table:diff_split}.

\begin{figure*}[t]
\begin{tabular}{ll ll}
\includegraphics[width=0.165\linewidth]{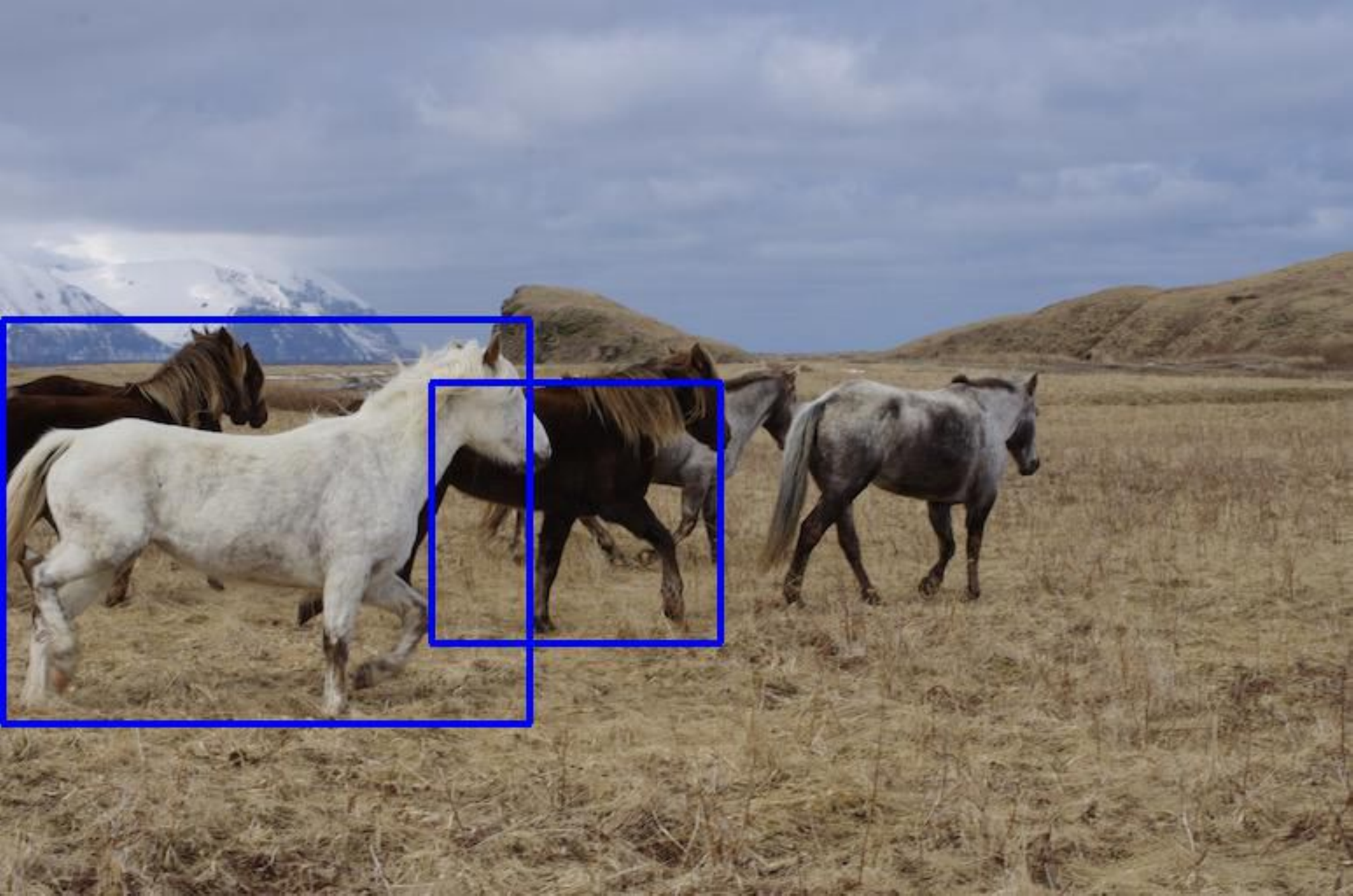}
\includegraphics[width=0.165\linewidth]{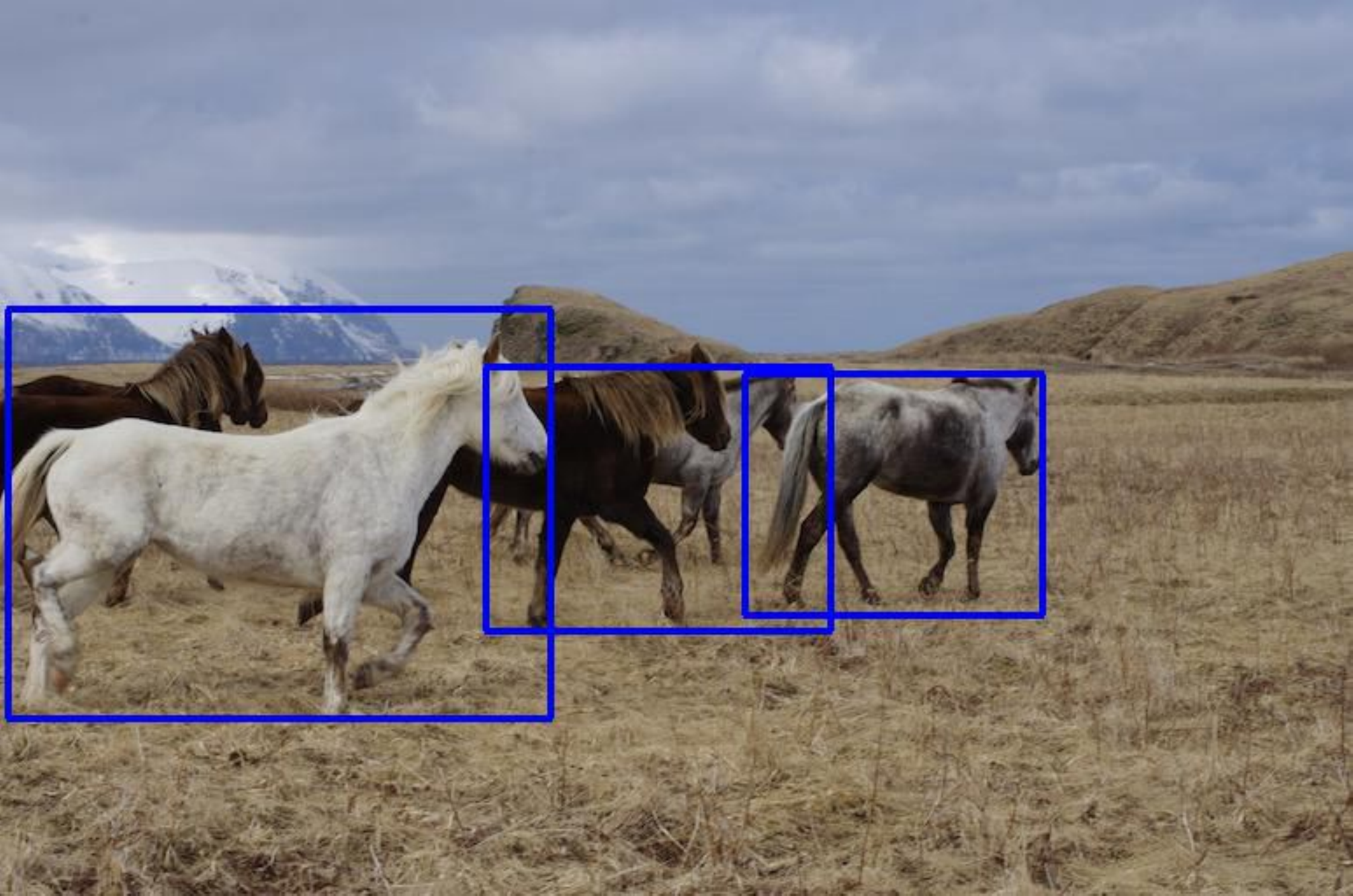}
\includegraphics[width=0.165\linewidth]{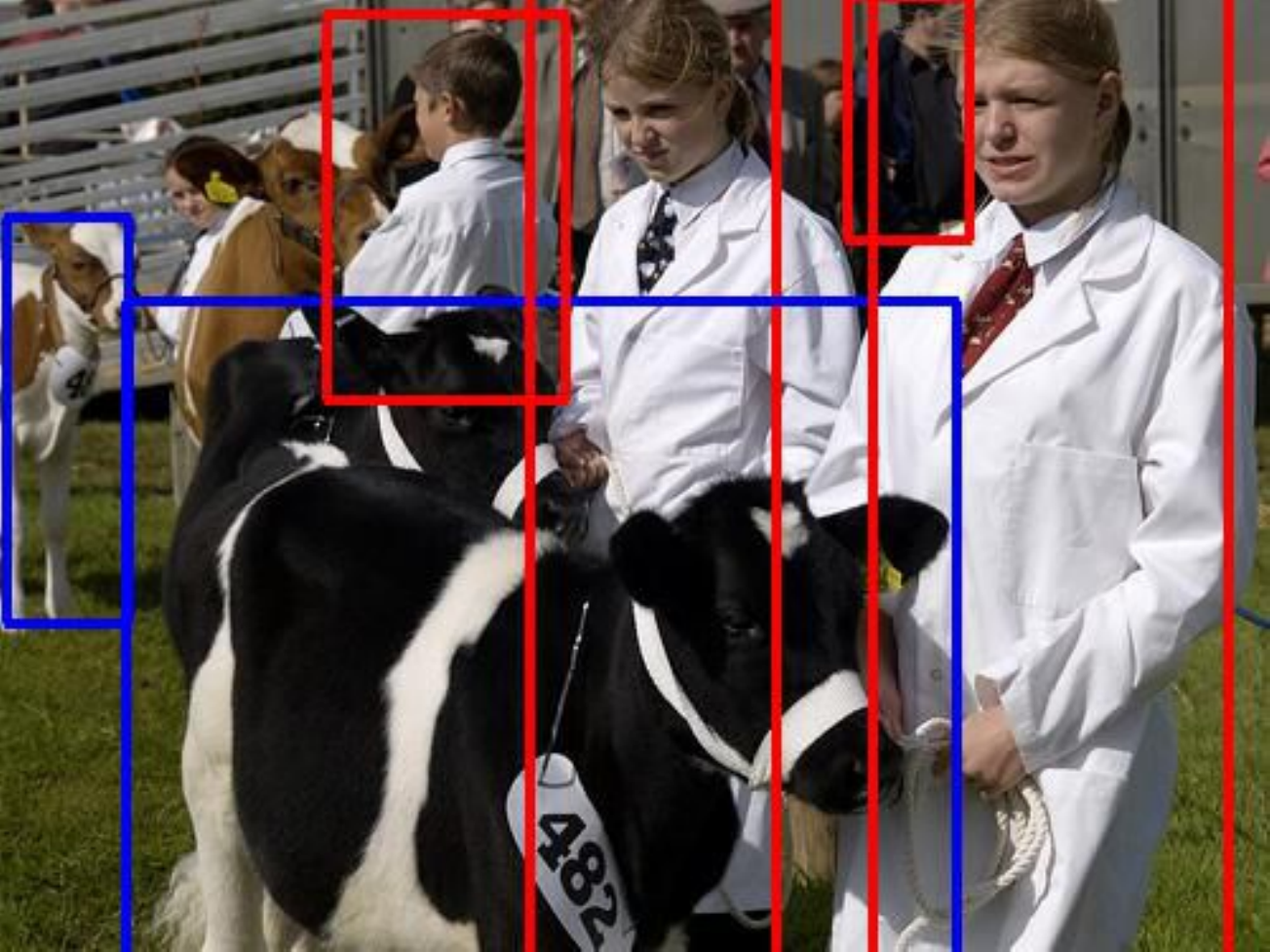}
\includegraphics[width=0.165\linewidth]{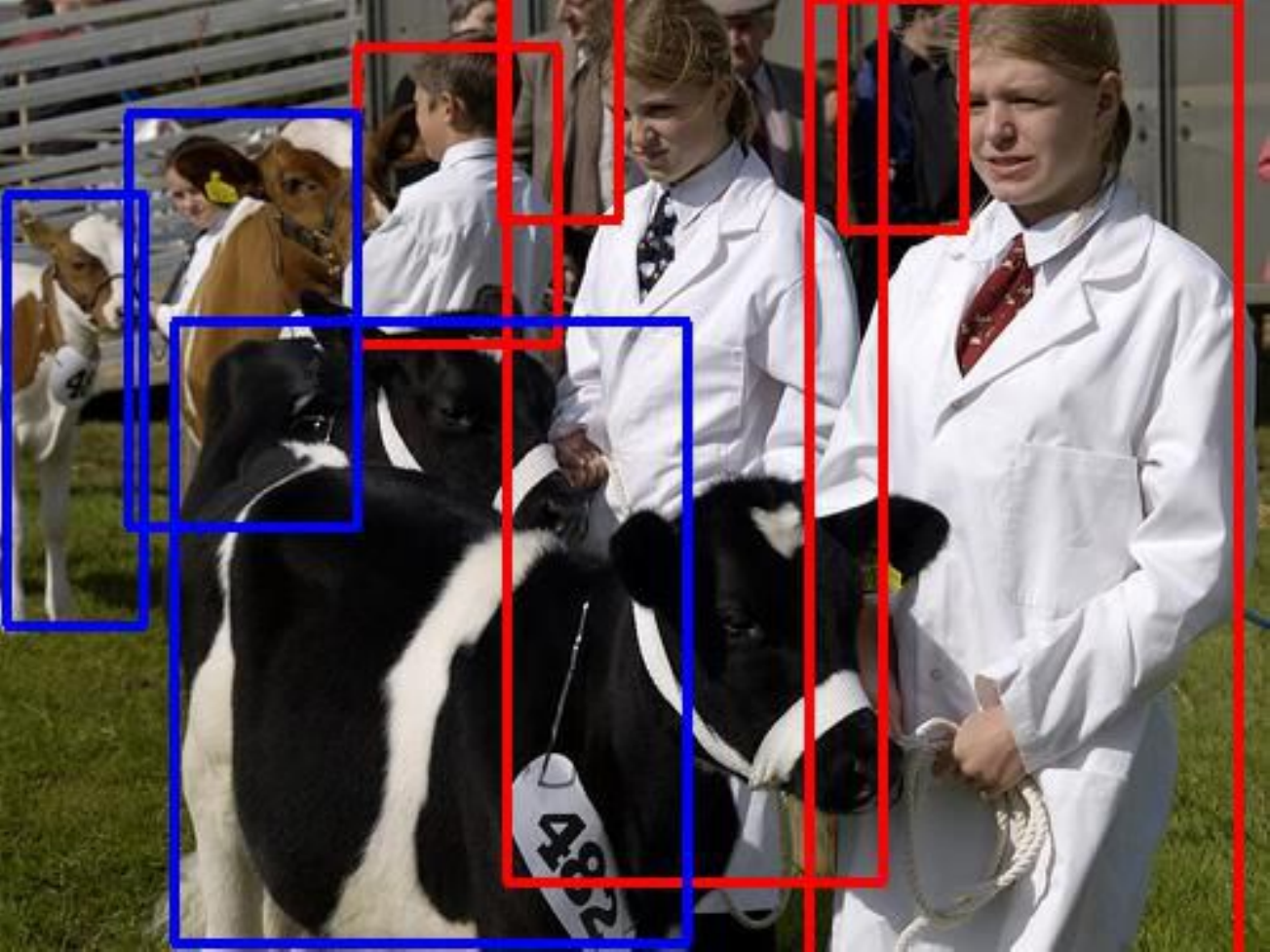}
\includegraphics[width=0.165\linewidth]{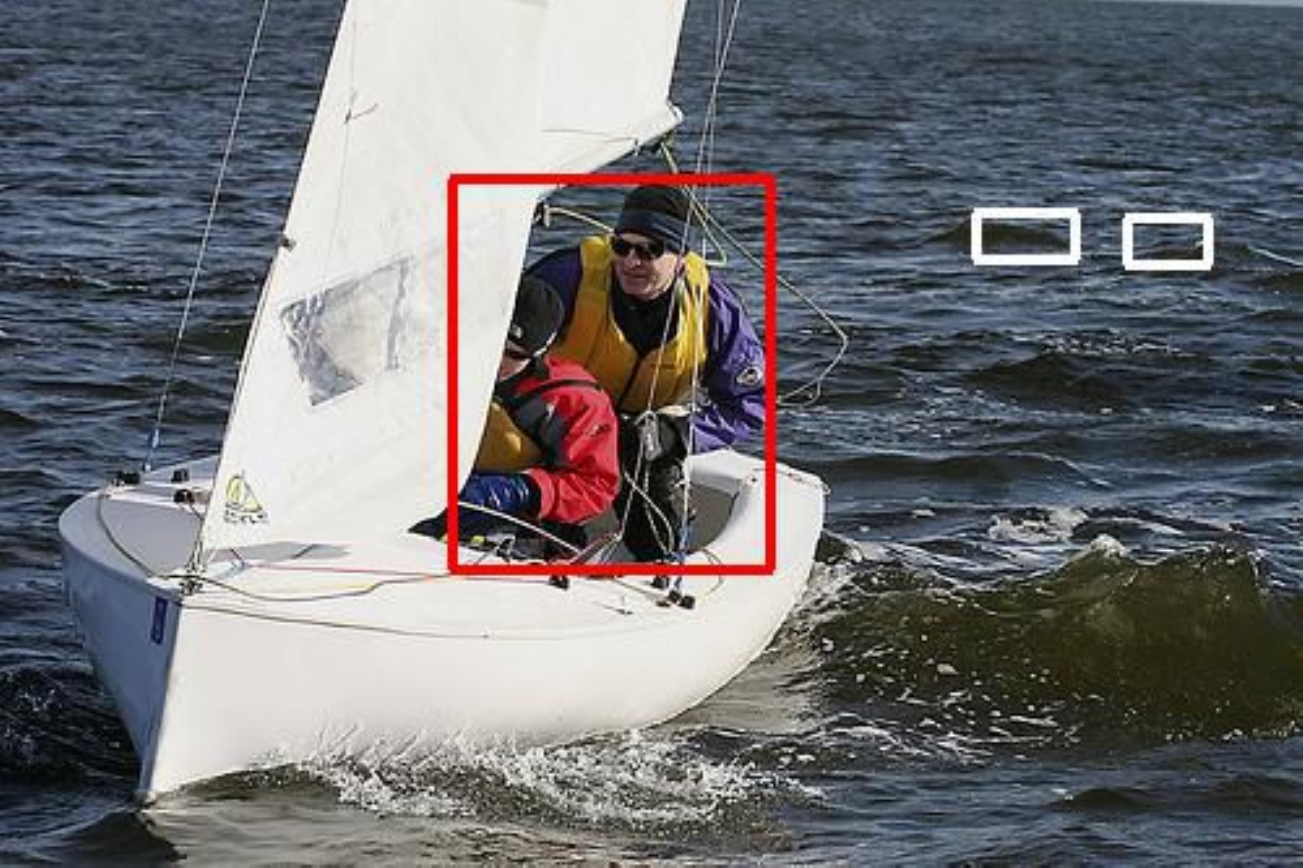}
\includegraphics[width=0.165\linewidth]{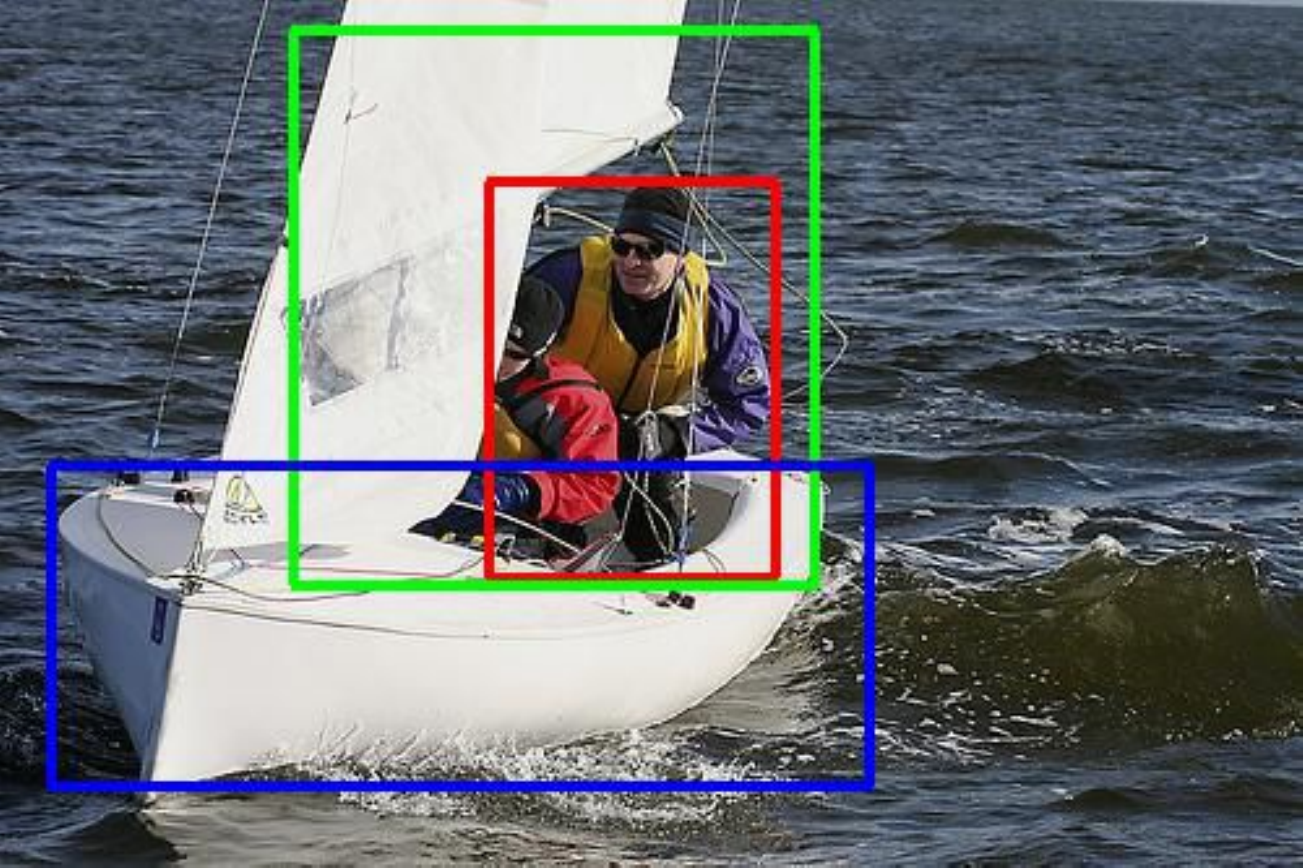}\\
\includegraphics[width=0.165\linewidth]{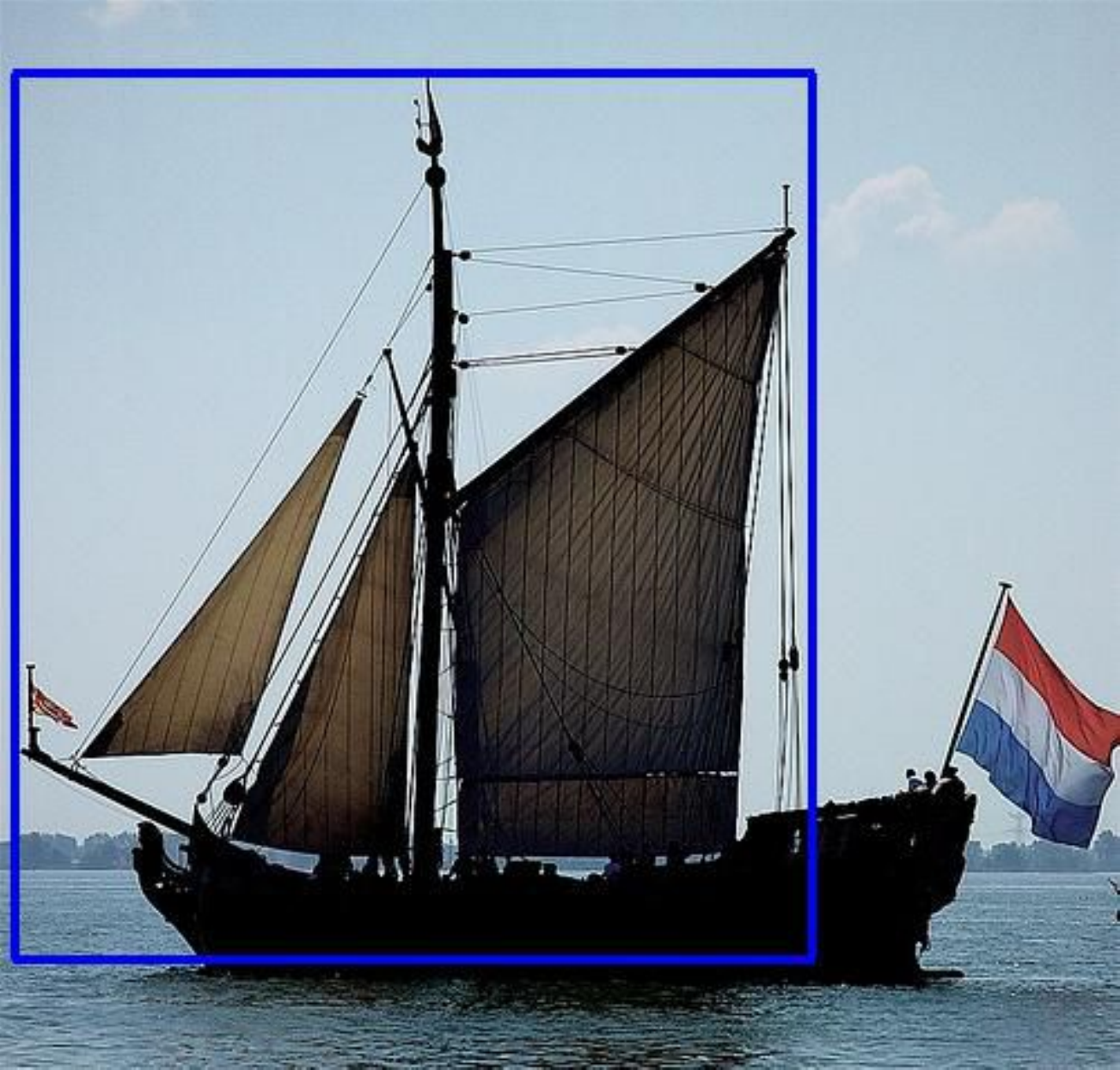}
\includegraphics[width=0.165\linewidth]{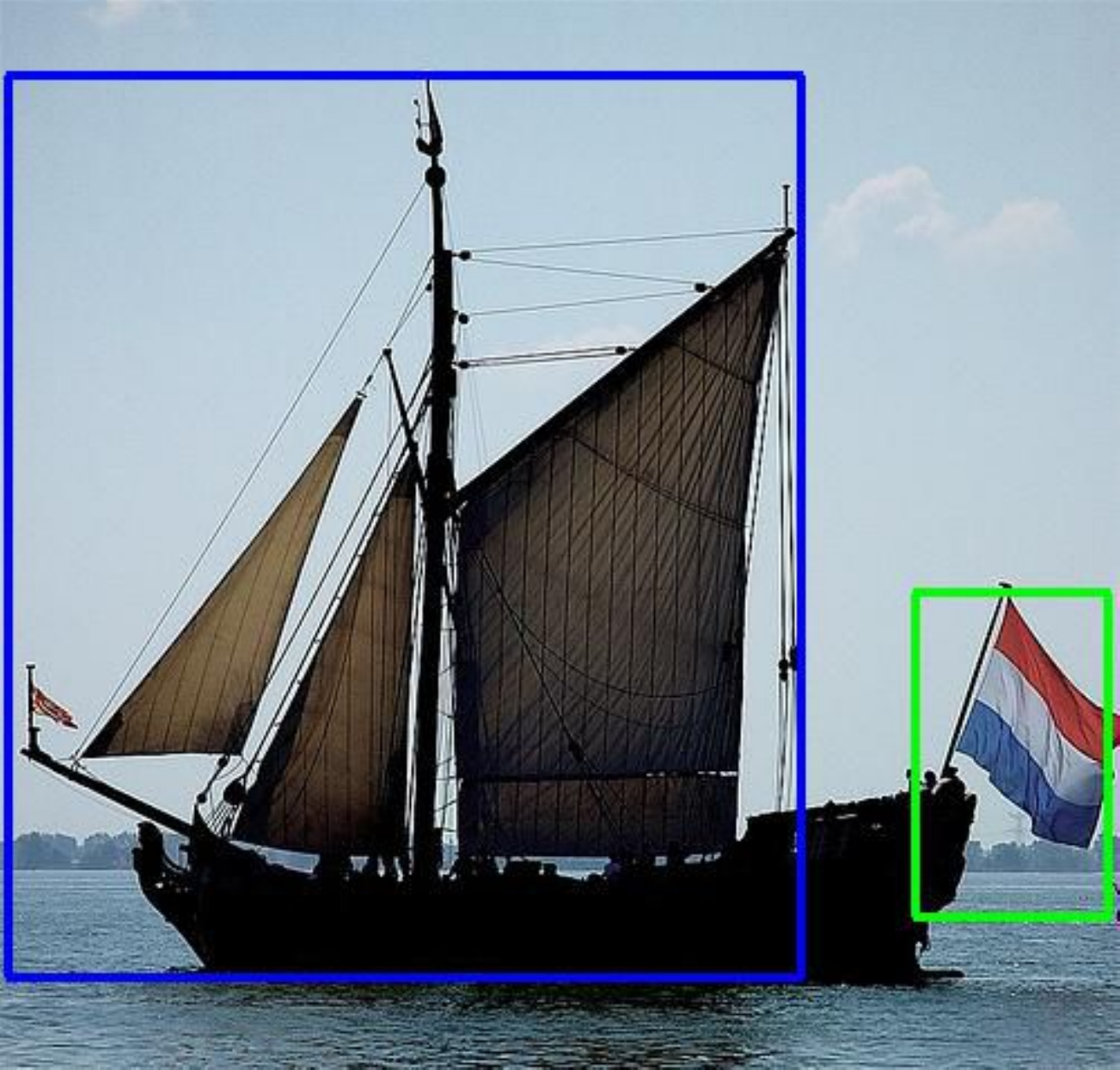}
\includegraphics[width=0.165\linewidth]{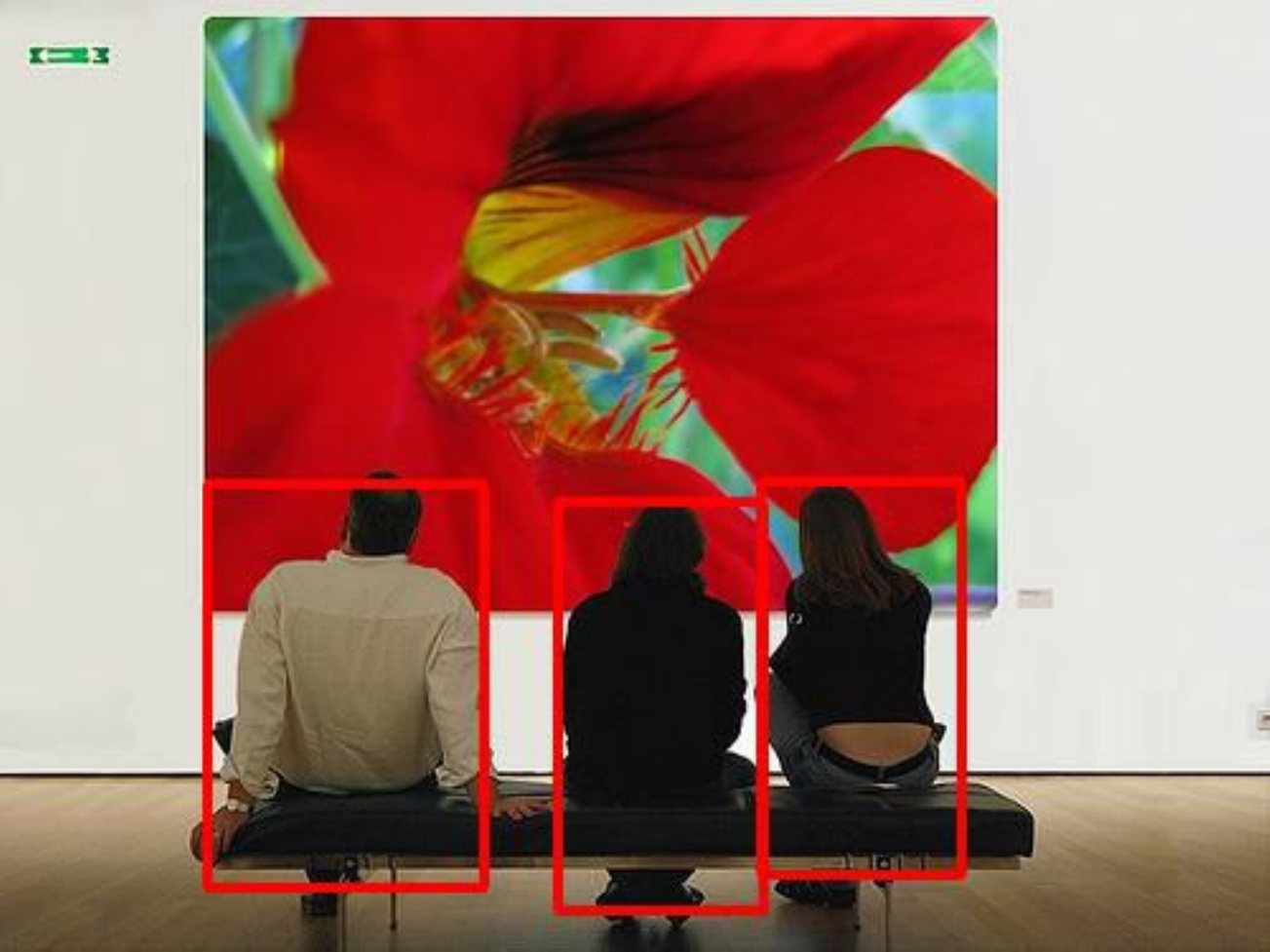}
\includegraphics[width=0.165\linewidth]{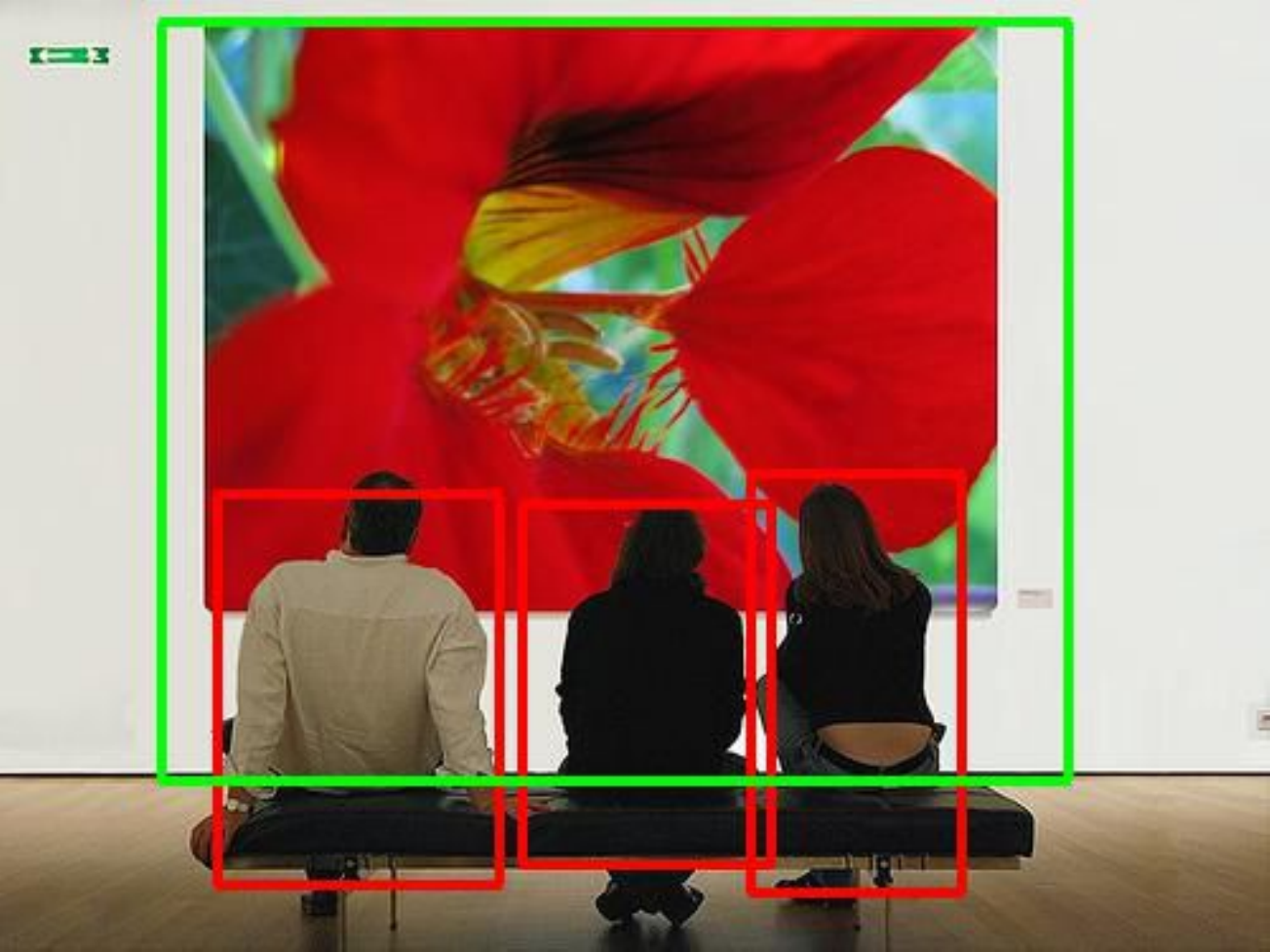}
\includegraphics[width=0.165\linewidth]{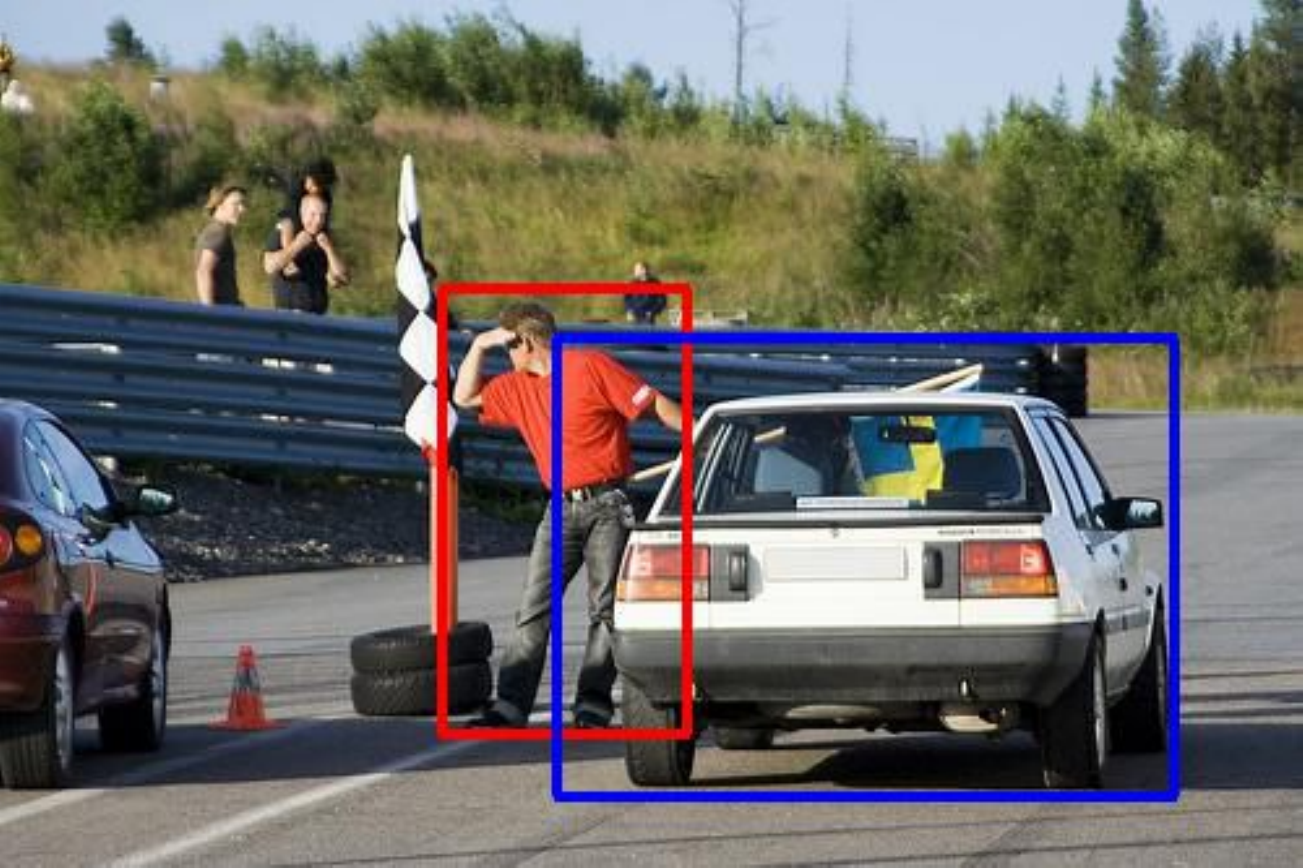}
\includegraphics[width=0.165\linewidth]{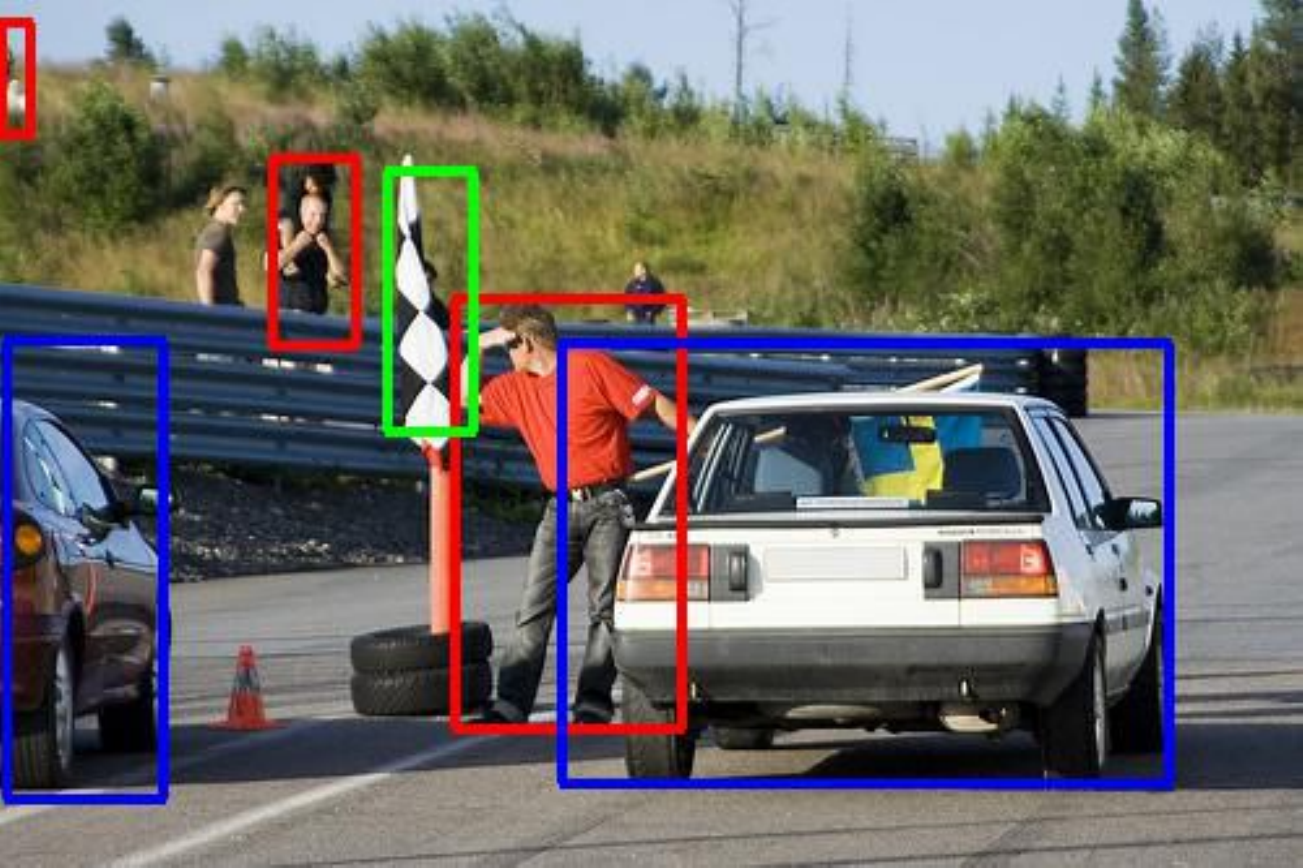}\\
\includegraphics[width=0.165\linewidth]{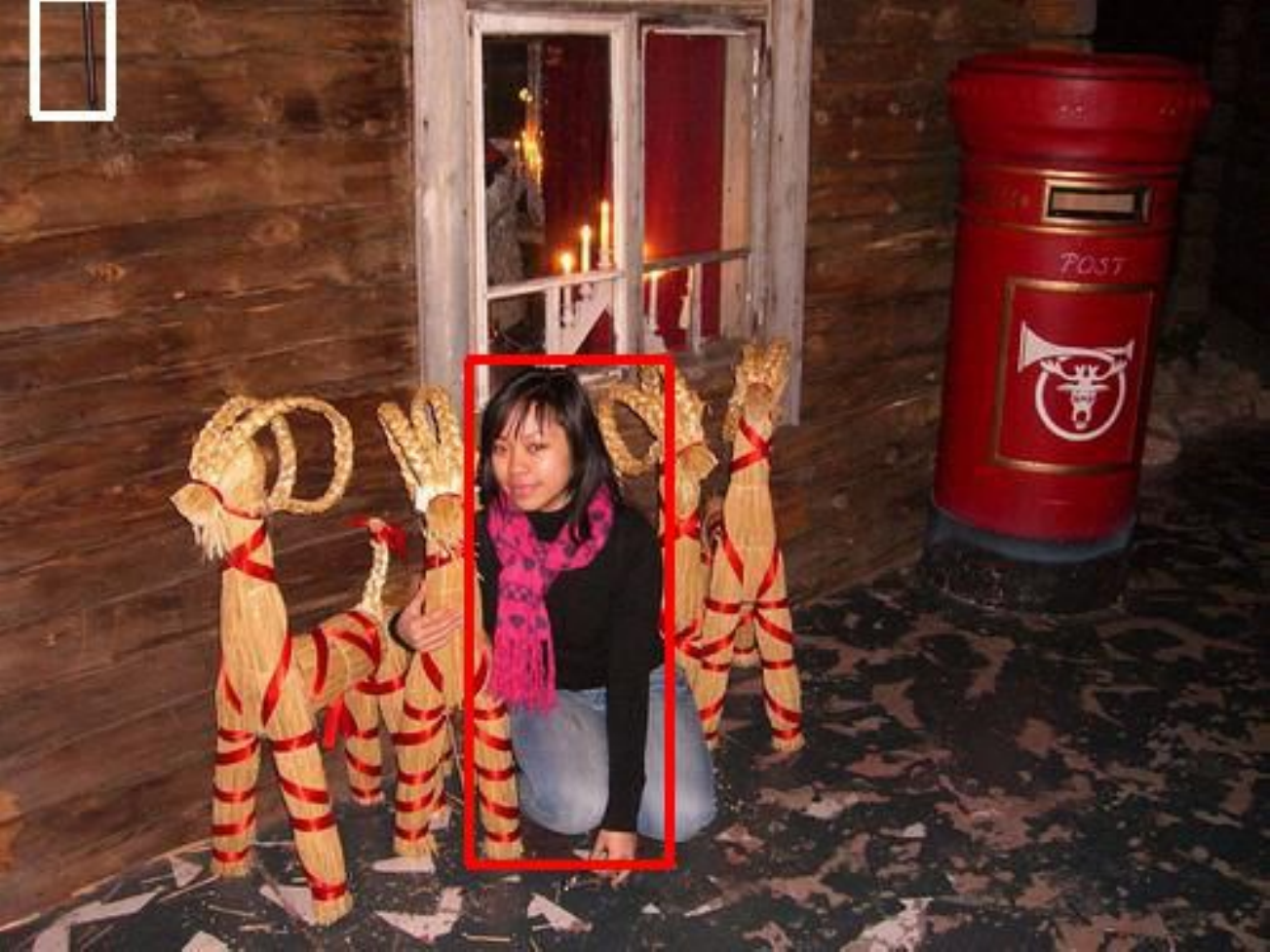}
\includegraphics[width=0.165\linewidth]{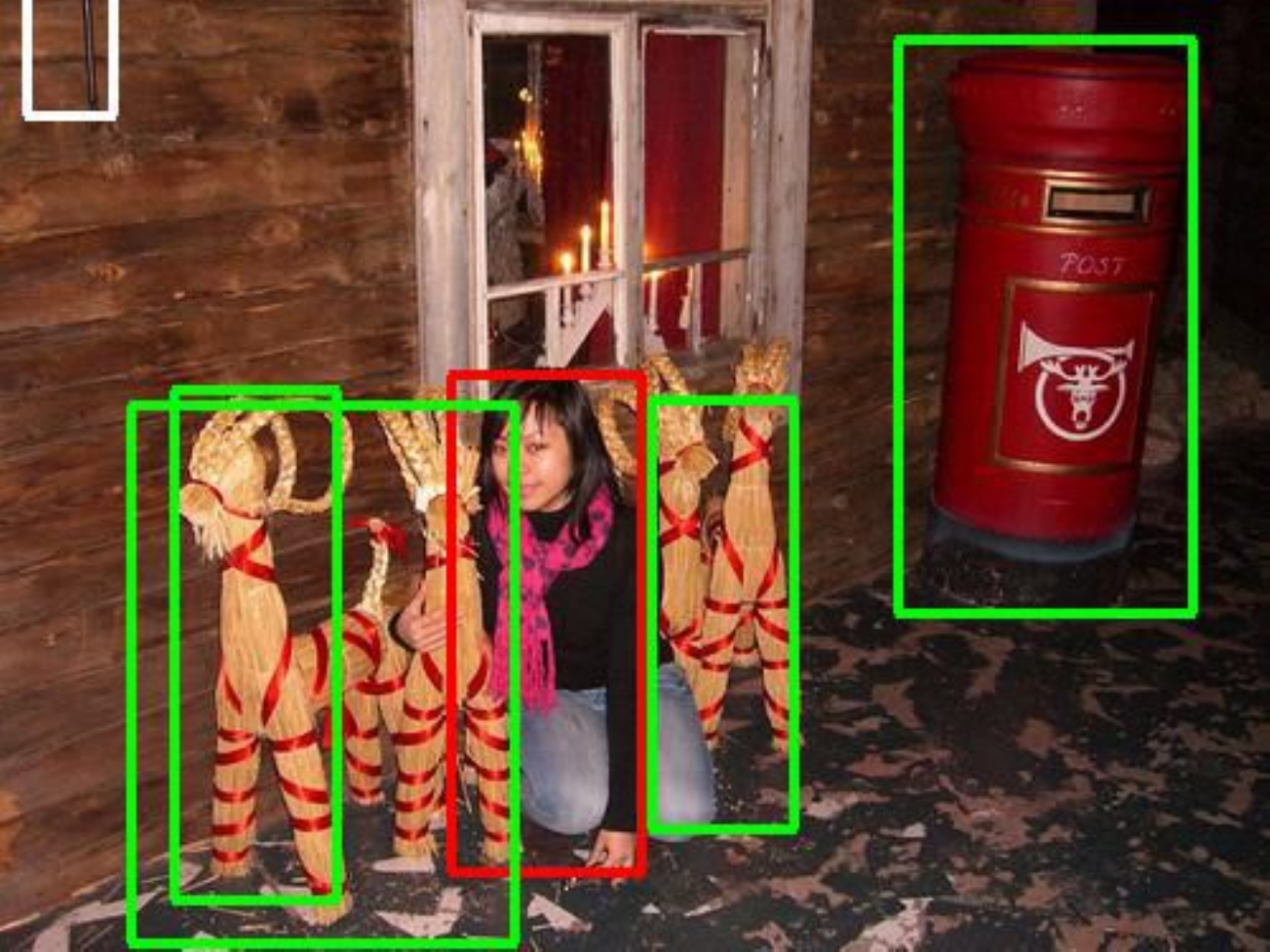}
\includegraphics[width=0.165\linewidth]{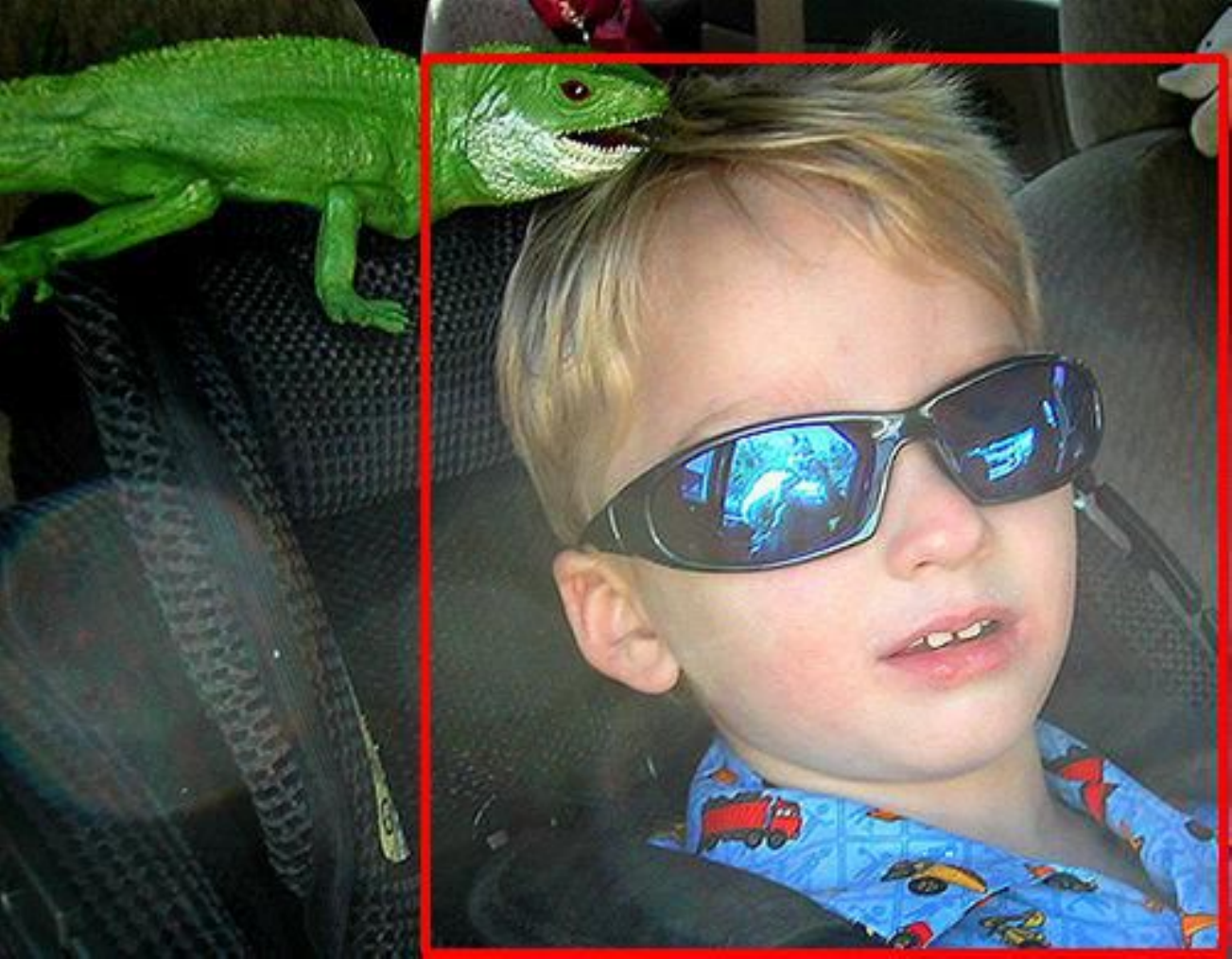}
\includegraphics[width=0.165\linewidth]{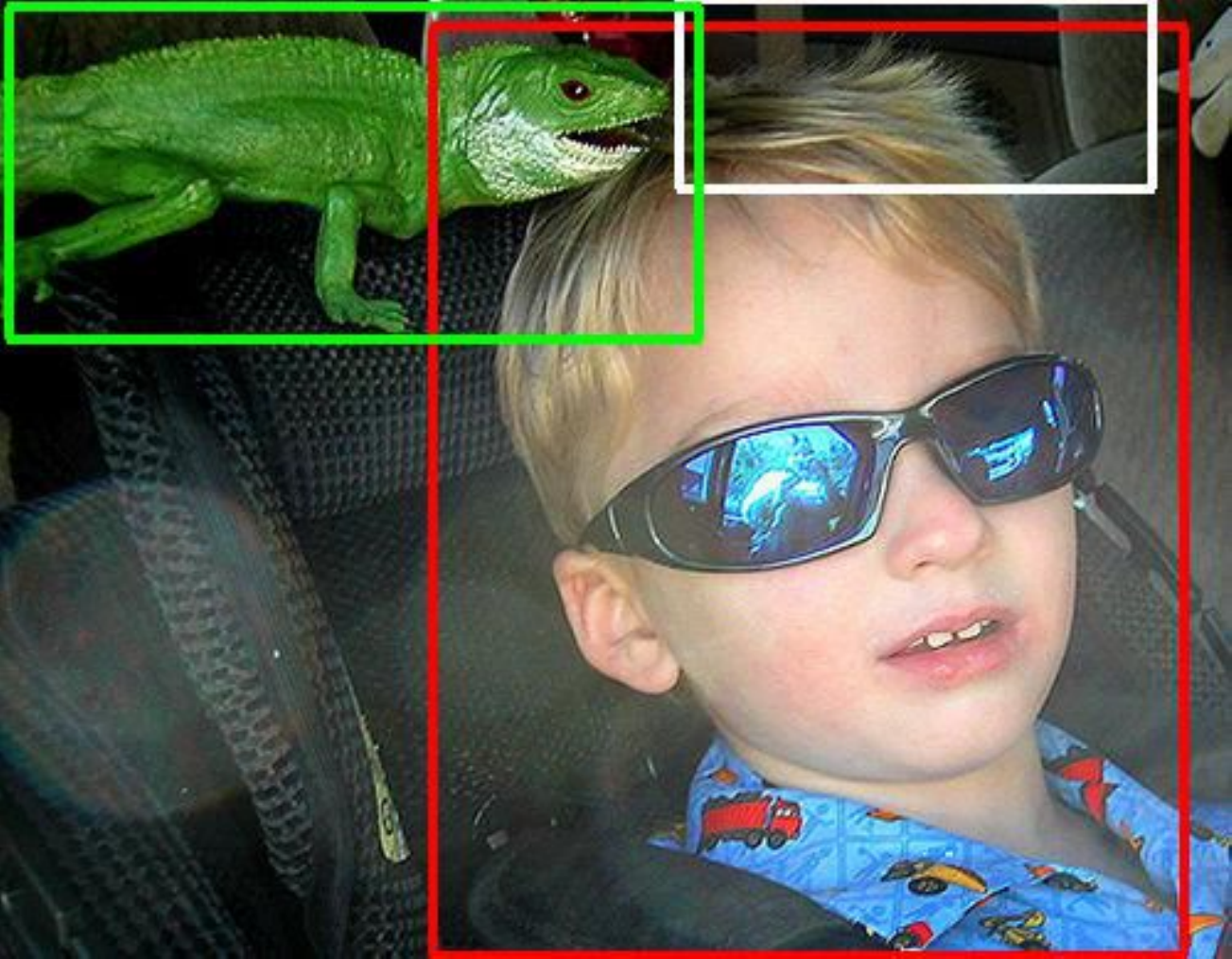}
\includegraphics[width=0.165\linewidth]{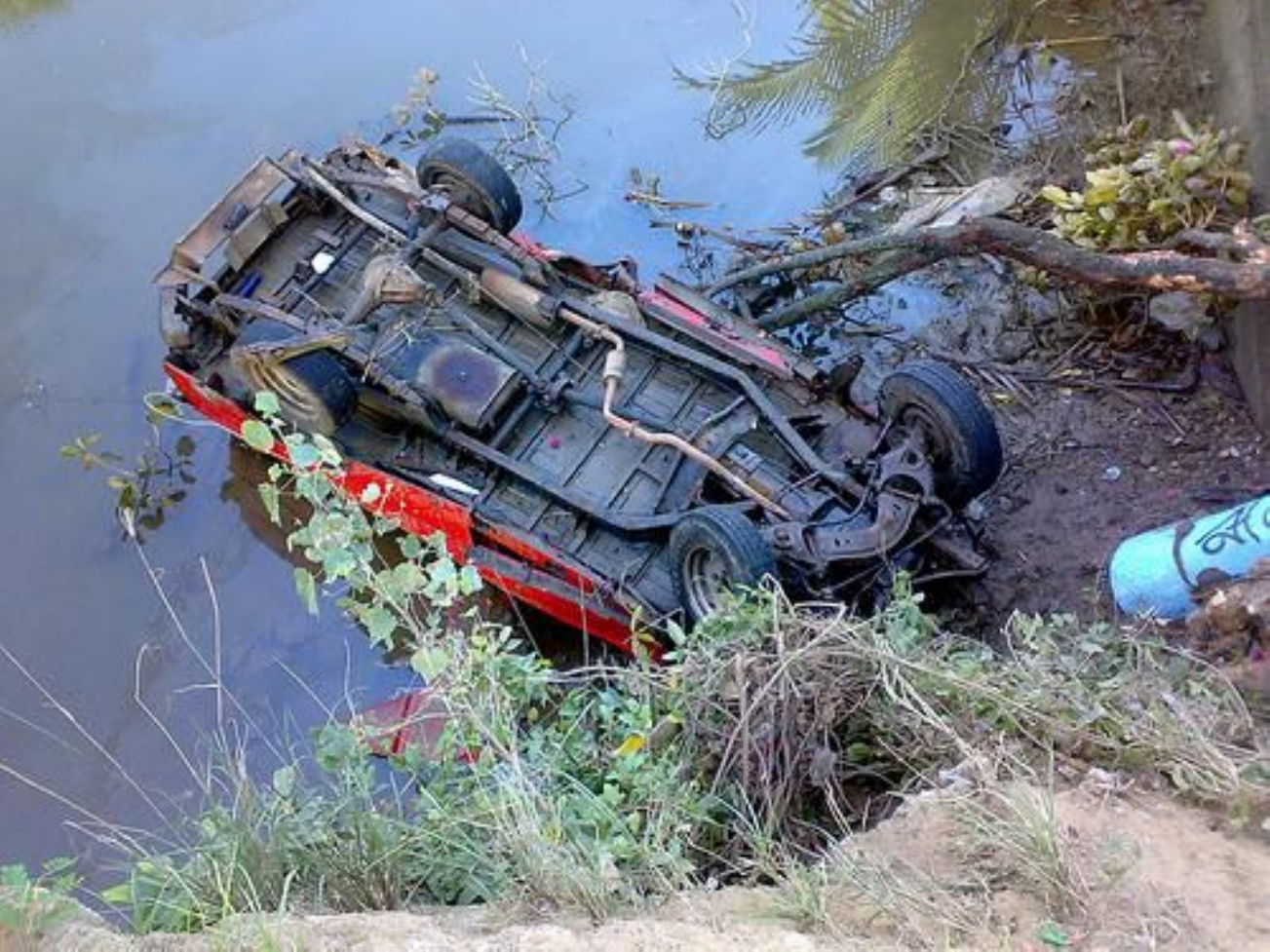}
\includegraphics[width=0.165\linewidth]{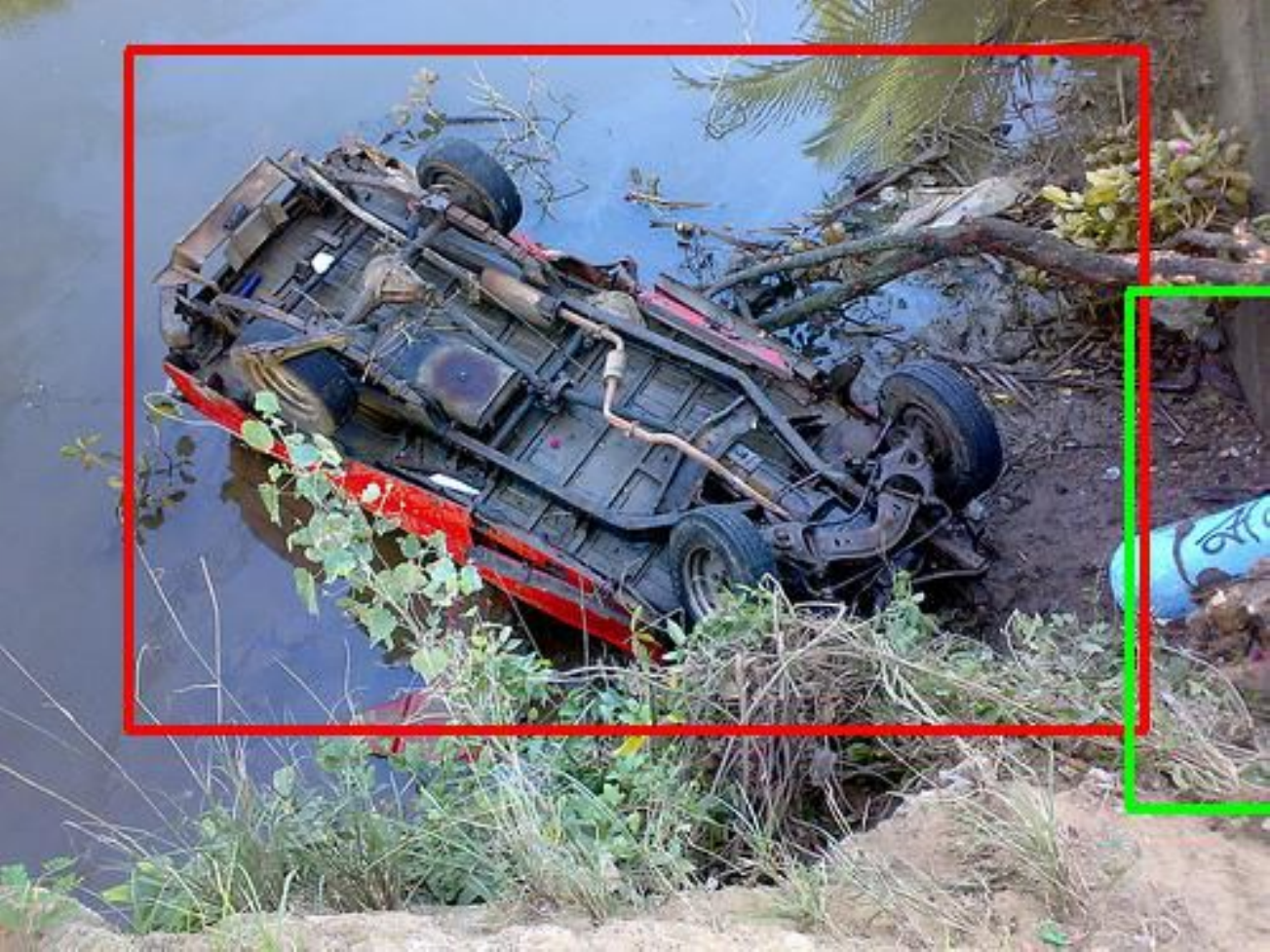}\\
\includegraphics[width=0.165\linewidth]{12_yolo.pdf}
\includegraphics[width=0.165\linewidth]{12_newarc.pdf}
\includegraphics[width=0.165\linewidth]{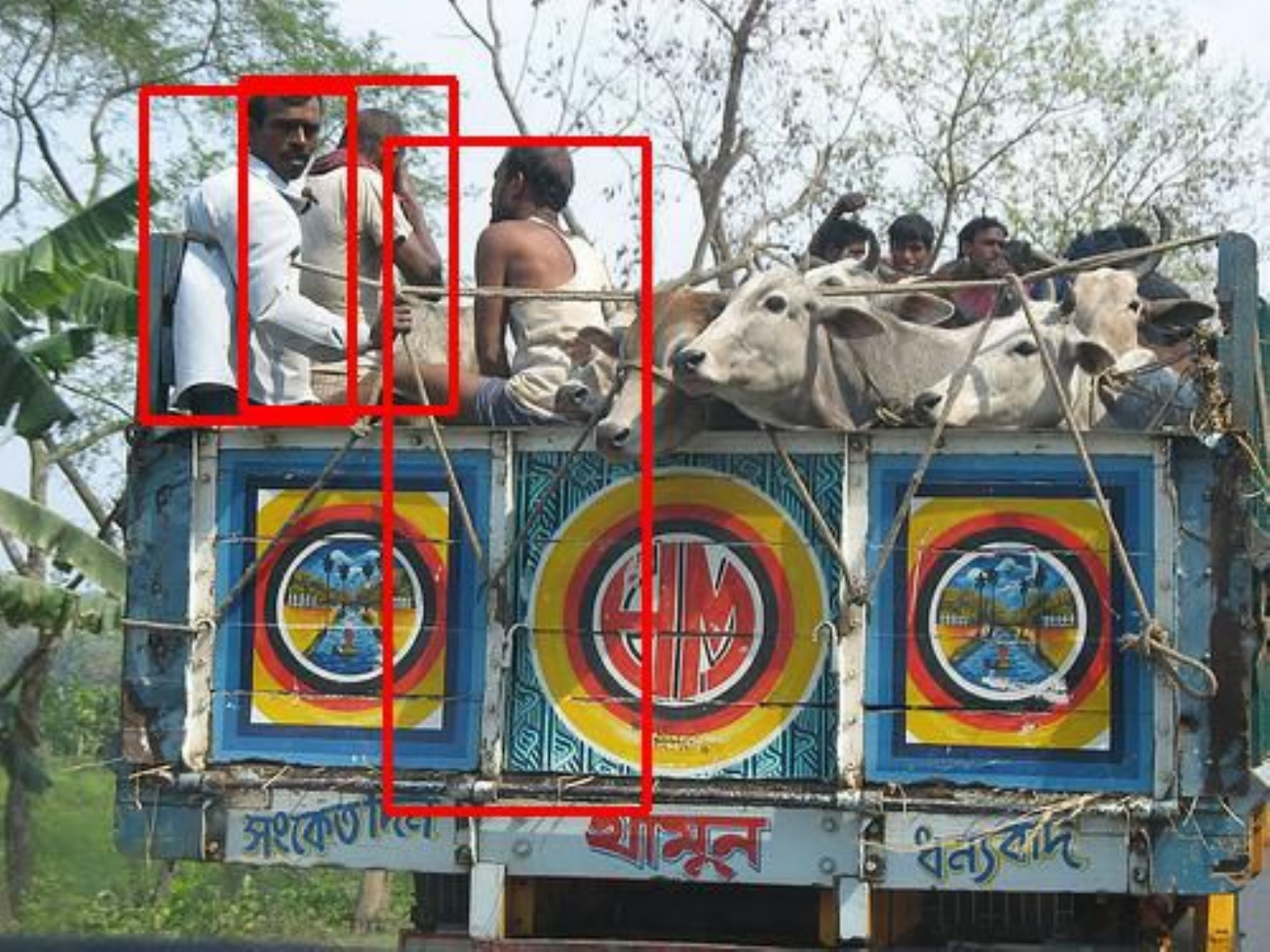}
\includegraphics[width=0.165\linewidth]{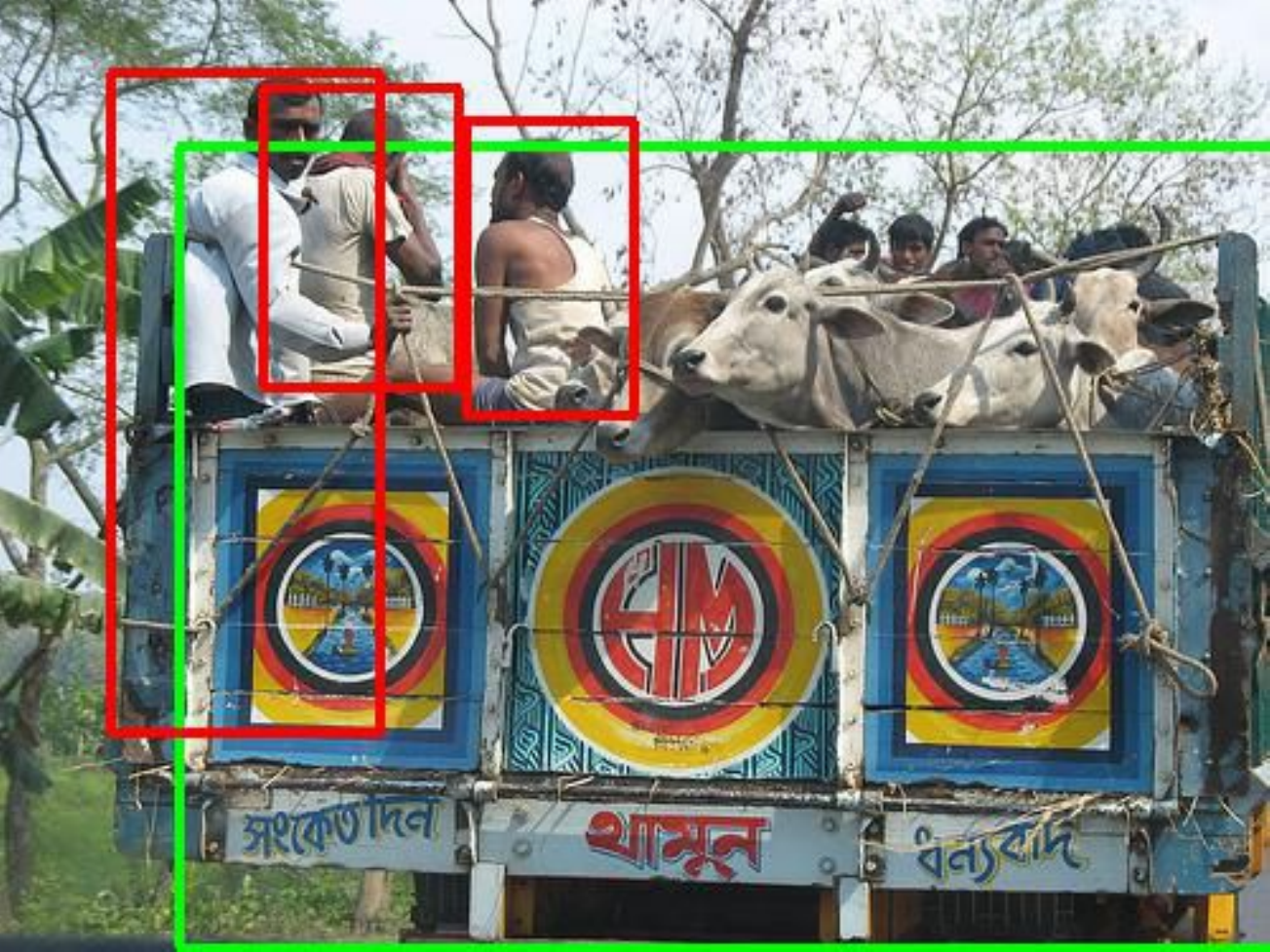}
\includegraphics[width=0.165\linewidth]{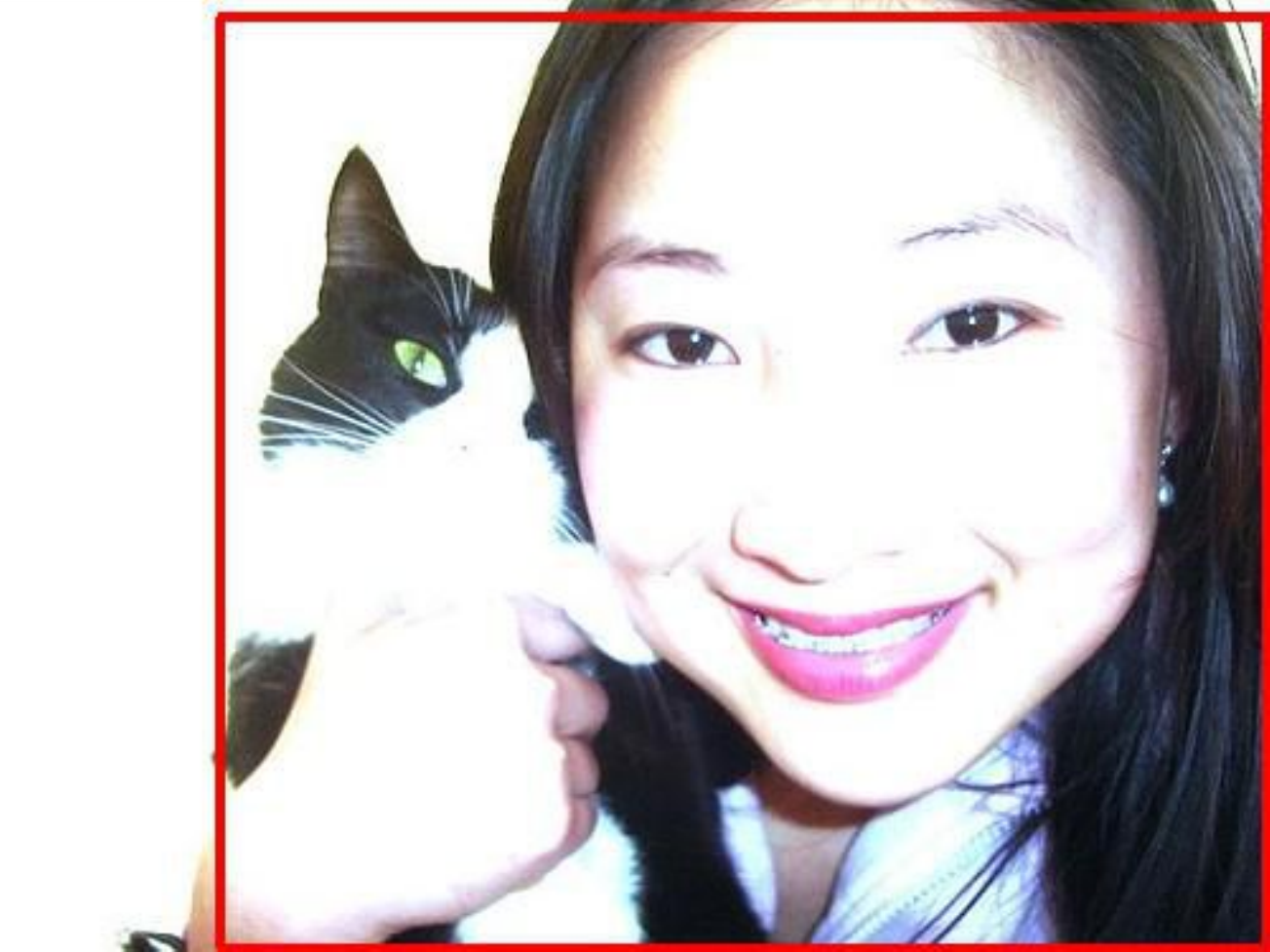}
\includegraphics[width=0.165\linewidth]{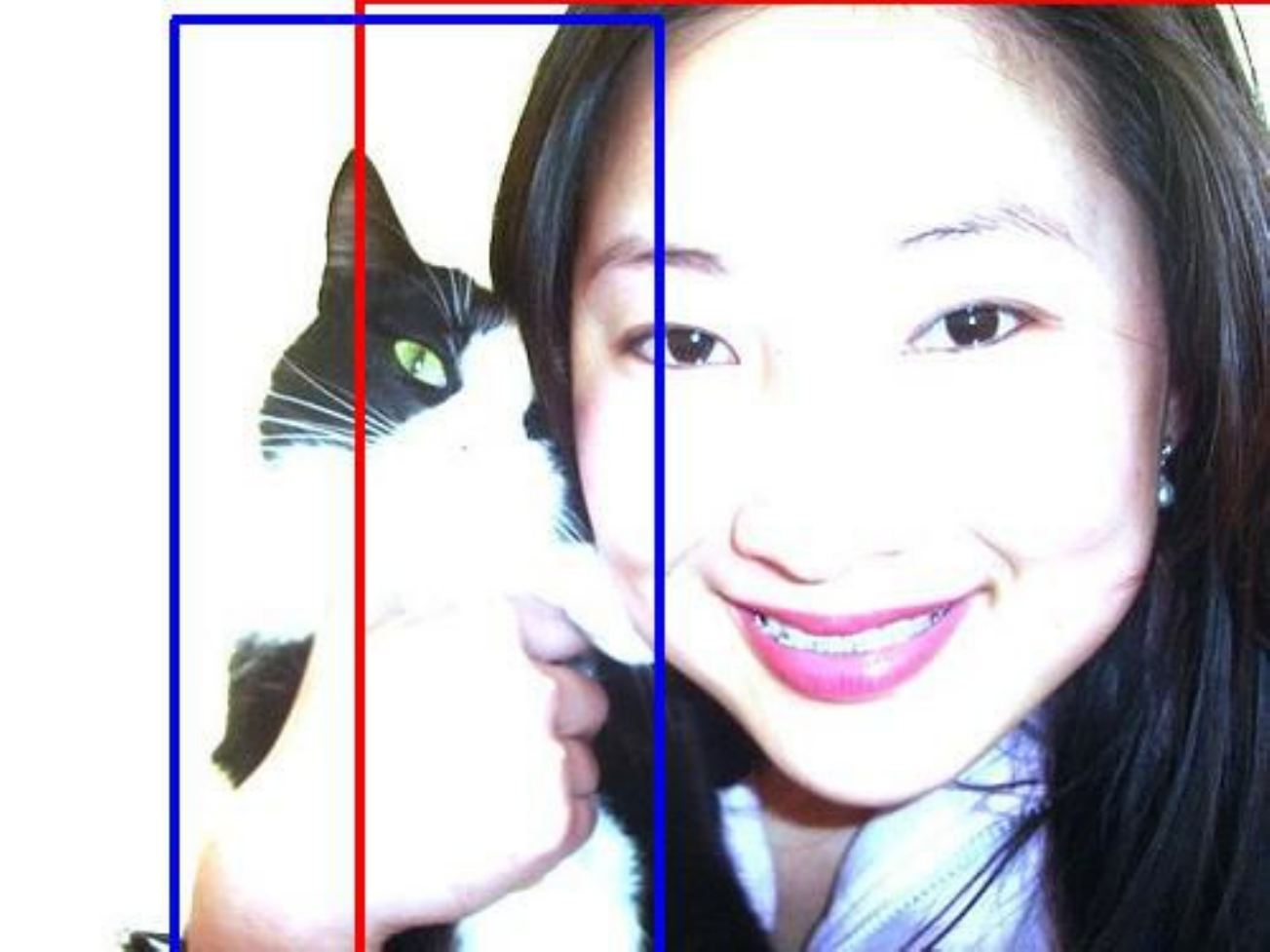}
\end{tabular}
\caption{\small (Best to be viewed in color) Examples of detected unseen classes by  YOLOv2 (left) and proposed ZS-YOLO (right). ZS-YOLO not only detects artificially held out unseen classes  such as sofa, sailboat, potted plant, cat, but also recognizes completely unannotated objects such as flag, painting, chameleon, pet horse, truck, etc. It is also interesting that car wreck is recognized by our model and missed by YOLOv2 although car is in the seen classes. We believe that training with semantic attributes prevent over-fitting of detector for specific view-angles as well. Although our model is better in overall performance, the lower overall precision can be partially attributed to unannotated object classes that count as false positives. The seen, unseen and unannotated classes are color coded as {\color{red}red}, {\color{blue}blue} and {\color{green}green}. \label{figure:image_comp}}
\end{figure*}

It might be surprising that the performance gap is not large over YOLOv2 on {\it PASCAL VOC} as revealed in TABLE~\ref{table:diff_split} and Fig.~\ref{fig:prec_recall}. 
Importantly, we argue that this is primarily because other than the unseen object classes among the 20 annotated ones we artificially held out during training, there are many more unannotated object classes in the {\it PASCAL VOC} dataset which do not have their corresponding ground truth boxes. Thus, they are also counted as false positives even though ZS-YOLO does successfully detect them as foregrounds. Unfortunately, these cases cannot be quantitatively measured. We thus show in 
Fig.~\ref{figure:image_comp} extensive qualitative results, where ZS-YOLO not only succeeds in detecting artificial unseen object classes, but also those that are truly unseen objects that have no annotation. These successful detections will be measured as false positives in TABLE~\ref{table:diff_split} and Fig.~\ref{fig:prec_recall}, resulting in lower AP. Therefore, ZS-YOLO's performance gain can be expected to be larger than YOLOv2 in practice.

\subsection{Ablative Study on PASCAL VOC}
\label{sec:analysis}

\begin{table}[t]
\centering
\setlength{\tabcolsep}{0.3cm}
\begin{tabular}{l c c c}
\hline
\multirow{2}{*}{Model} & \multicolumn{3}{c}{AP}\\
 & Unseen & Seen & Mix \\
\hline
YOLOv2 & 56.4 & {\bf 71.6} & {\bf 54.3} \\
ZS-YOLO(visual) & 57.2 & 71.2 & 52.3 \\
ZS-YOLO(semantic) & 57.0 & 70.9 & 52.3 \\
ZS-YOLO(full) & \bf{60.1} & 71.0 & 53.9 \\
\hline \\
\end{tabular}
\caption{\small Ablative analysis of different components of the proposed ZS-YOLO model. }
\label{table:ablative}
\end{table}

\begin{table}[t]
\centering
\renewcommand{\arraystretch}{1.2}
\setlength{\tabcolsep}{0.2cm}
\begin{tabular}{l | c c c c}
\hline
Prototypes & attribute & w2v & one-hot & random\\
Unseen AP & {\bf 60.1} & 58.9 & 57.7 & 49.0 \\
\hline
\multicolumn{3}{c}{}\\
\end{tabular}
\caption{Semantic prototype comparison on {\it PASCAL VOC} 10/10 split. \label{table:diff_attr}}
\end{table}

\subsubsection{Effect of Semantic Prediction}
To further evaluate the effect of the semantic prediction side task during model training and its impact on the learnt visual features, we conducted ablative analysis on the proposed ZS-YOLO model. Specifically, we trained two different models for the {\it PASCAL VOC} 10/10 split: (1) ZS-YOLO (visual): Instead of using the combined multi-model input $\left[\mathbf{T}_F, \mathbf{T}_L, \mathbf{T}_S\right]$ (see Fig.~\ref{figure:arc}) for final confidence score prediction, we removed the semantic prediction module completely from the network architecture, so the final confidence prediction is based purely on $\left[\mathbf{T}_F, \mathbf{T}_L\right]$ without semantic information; and (2) ZS-YOLO (semantic): In this model, we remove visual features from confidence prediction input by only feeding $\left[\mathbf{T}_{L}, \mathbf{T}_{S}\right]$ into the confidence prediction module. Other than the above two models, we consider original YOLOv2 which only uses $\mathbf{T}_F$ for confidence prediction as a third baseline. The comparative results are shown in TABLE~\ref{table:ablative}.

\begin{figure}[t]
\centering
\includegraphics[width=0.49\linewidth]{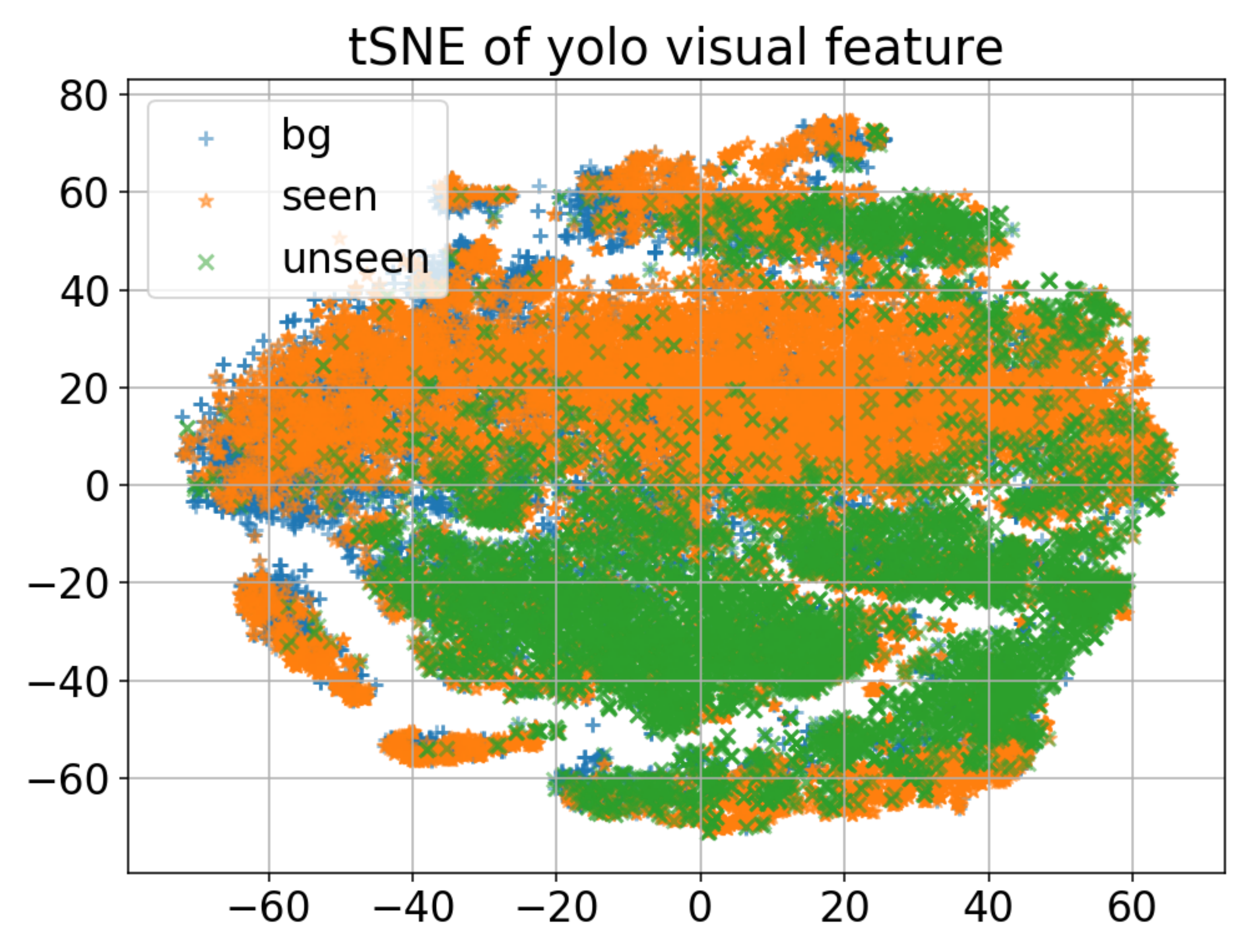}
\includegraphics[width=0.49\linewidth]{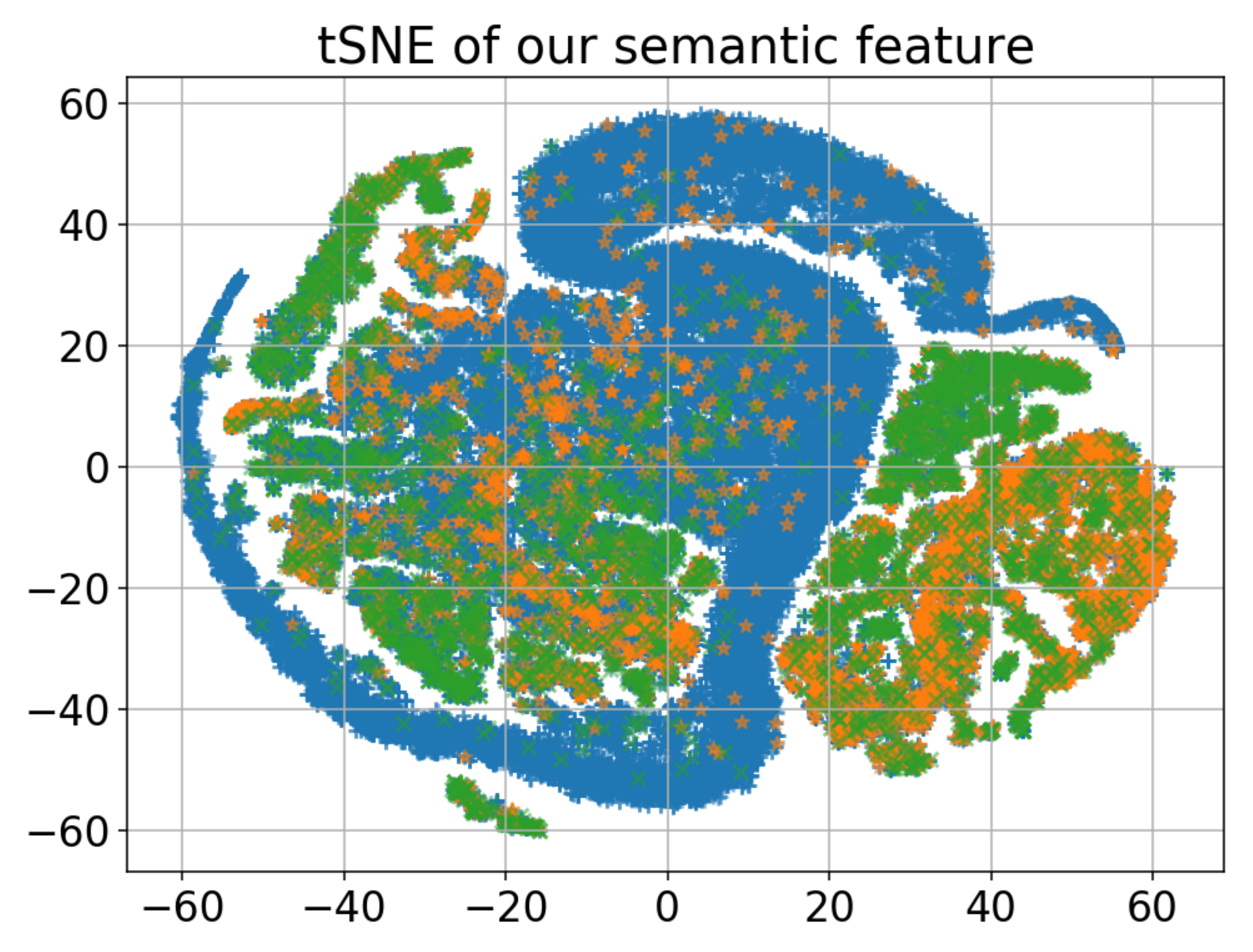}
\captionof{figure}{\small tSNE visualization comparison between YOLOv2 visual features (left) with ZS-YOLO semantic features (right) on Test-Mix dataset. Foreground ({\color{orange} seen} and {\color{green} unseen}) and background ({\color{blue}bg}) are more separable if trained with the semantic prediction as a side task.  \label{figure:tSNE}}
\end{figure}

From TABLE~\ref{table:ablative} we observe that utilizing both visual features as well as semantic attributes as multi-modal clues improves the detection performances on unseen data, compared to detection models that exclusively use single-modal information to infer objectness confidence. Specifically, on unseen data the ZS-YOLO(full) model that explores both visual and semantic domains achieves $60.1\%$ AP on Test-Unseen data, which is the highest among all three competitors. By comparing YOLOv2 with ZS-YOLO(visual), we can see that using the bounding box locations as a side information also helps to detect unseen objects, although such information might be redundant for detecting seen objects especially when the visual features are uniquely fine-tuned to those classes.
Finally, we can see when we exclusively utilize only semantic information, ZS-YOLO(semantic) generates better detection performance compared to YOLOv2 on unseen data, which again validates the importance of utilizing semantic information when moving toward zero-shot detection. 
This effect can be visualized in Fig.~\ref{figure:tSNE} with t-SNE embeddings of all the bounding box proposals generated on the {\it PASCAL VOC} Test-Mix data by using the learned YOLOv2's visual feature as well as ZS-YOLO's semantic feature (i.e. $\mathbf{T}_S$). 

\begin{table}[t]
\centering
\setlength{\tabcolsep}{0.2cm}
\begin{tabular}{l c c c c}
\hline
{Dataset} & YOLOv2 & Ours({$\mathbf{Atr}$}) & Ours(w2v) & Ours(w2vR)\\
\hline
\hline
PASCAL VOC &56.4  & \bf{60.1} & 58.9 & 59.3\\
MS COCO & 37.3 & 38.4 & 38.4 & \bf{40.6}\\
\hline \\
\end{tabular}
\caption{AP \% for unseen classes with different semantic features on {\it PASCAL VOC}  (10/10 split) and {\it MS COCO} (20/20 split).\label{table:diff_semantic}}
\label{table:1}
\end{table}

\begin{table}[t]
\centering
\setlength{\tabcolsep}{0.3cm}
\begin{tabular}{l c c c}
\hline
Split & $E$ Score & AP &\\
\hline
10/10-{\bf1} & {\bf 0.843} & {\bf 60.1} &\\
10/10-{\bf2} & 0.793 & 58.7 &\\
10/10-{\bf3} & 0.567 & 39.3 &\\
\hline
5/15-{\bf1} & 0.684 & 37.3 &\\
5/15-{\bf2} & 0.630 & 34.7 &\\
\hline \\
\end{tabular}
\caption{\small Seen/Unseen correlation comparison on {\it PASCAL VOC}. Unseen performance correlates strongly with $E$ score. \label{table:attribute_corr}}
\end{table}

\subsubsection{Choices of Semantic Prototypes: One-hot \& Random}
We are also interested in the problem of whether using attributes and word2vec (w2v) as semantic prototypes are indeed superior to other choices. Two semantic prototype spaces were tested for training in this context: (1) One-hot encoding: we use the one-hot class label vectors as semantic prototypes, so that all the prototypes are orthogonal to each other. Note that one-hot vectors were used only to encode seen classes. Since, we disconnect semantic attribute predictor at test-time, the issue of unseen semantic representation does not arise for our situation. Indeed, were we to look at semantic predictions, it is quite likely that at test-time the unseen classes are possibly represented by a combination of seen (one-hot) classes, analogous to the semantic attributes. (2) Random encoding: We generate the prototypes for each class with same dimension as attribute vectors ($h=64$) by randomly sampling from uniform distribution. We run the experiments on the {\it PASCAL VOC} 10/10 split and the comparative results are reported in TABLE~\ref{table:diff_attr}. The results clearly show that by using meaningful semantic prototypes such as attributes and word2vec, the detection performance on unseen objects can be improved. Random encoding prototypes, which contain no semantic information, make the AP of our model 49.0\%, significantly worse than original YOLOv2. Meanwhile, by utilizing one-hot encoding semantic prototypes, class level information is introduced and our model can get better detection performance on unseen data.

\subsubsection{Effect of Semantic Dimensionality Reduction}
To verify our semantic dimensionality reduction method as discussed in Section \ref{sec:setting}, we compared models trained in the reduced word2vec space (w2vR) with the original word2vec space (w2v) on both {\it PASCAL VOC} (10/10 split) and {\it MS COCO} (20/20 split) datasets. Since the original w2v space (300 dimensions) is highly noisy, which is irrelevant to the visual space, learning such information has the effect of poor convergence behavior for a convolutional neural network. On the contrary, the reduced word2vec space w2vR is defined to preserve similarity structures originally defined in the attribute space which is more visually related. It is evident from TABLE~\ref{table:diff_semantic} that models trained on our mapped w2v, w2vR, leads to better performance on detecting unseen objects than the original w2v. In particular, on {\it MS COCO} dataset, the model trained on w2vR gets an AP of 40.6\%, even higher than model trained on VOC attribute, 38.4\%. 

\subsubsection{Effect of Seen/Unseen Correlations}
{In zero shot recognition, the unseen and seen classes have to share some common visual appearance such that the classifier can generalize the representation learned from seen classes to unseen\cite{zhang2015zero}. If the unseen classes have low correlation with seen classes, the zero shot task tends to be more difficult. The same phenomenon may also exist in zero shot detection. Therefore,}
we seek to test the effect of correlation between the attributes of seen and unseen classes on detection performance of unseen objects. To measure this correlation, we define an energy score function $E$ of a class split as:
$
E = \frac{1}{|\mathbf{C}_{unseen}|}\sum_{a\in\mathbf{C}_{unseen}}\max_{b\in\mathbf{C}_{seen}} S(\mathbf{y}_a, \mathbf{y}_b)
$.

If the energy score is higher, the unseen classes have higher correlation with the seen classes. To further analyze its effect, we construct two more 10/10 splits (named as 10/10-{\bf2} and 10/10-{\bf3}) and one 5/15 (5/15-{\bf2}) split experiments over {\it PASCAL VOC}.
{The splits are constructed based on the energy score. 10/10-2 has an intermediate energy score over all 10/10 splits while 10/10-3 has the smallest. 5/15-2 has the smallest energy score and since it is very close to 5/15-1 which already has the highest energy score we don't need to construct one more. When the energy score is high, visually similar classes seem to appear in seen/unseen split separately, and when it is low, they seem both in seen or unseen. For example, in 10/10-1 split, for some similar class pairs like motorbike-bicycle, bicycle is in seen and motorbike is in unseen. While in 10/10-3 split, they are both in seen. \footnote{{The detailed data splits are available at the first author's github page.}}}
Our results are reported in TABLE~\ref{table:attribute_corr},  with each split's seen/unseen correlation calculated by $E$ score defined in Section \ref{sec:setting}.
TABLE~\ref{table:attribute_corr} is consistent with our intuition, that in general when seen and unseen data are more semantically correlated, our ZS-YOLO performs better on detecting unseen classes. For instance, in split 10/10-{\bf 1} where $E=0.843$, ZS-YOLO achieves a $60.1\%$ AP on Test-Unseen; Whilst on split 10/10-{\bf 3} where seen/unseen classes are less semantically related ($E = 0.567$), ZS-YOLO achieves a inferior AP of $39.3\%$.

\begin{table*}[t]
\centering
\renewcommand{\arraystretch}{1.2}
\footnotesize
\setlength{\tabcolsep}{0.04cm}
\begin{tabular}{c | l c c c c c c c c c c | c c c c c c c c c c | c c c}
\hline
& Method & bike & bird & btl & bus & chair & dtbl & dog & prsn & pplnt & tv & apln & boat & car & cat & cow & horse & mtbk & sheep & sofa & train & seen & unseen & mAP\\
\hline
\hline
\multirow{2}{*}{\rotatebox{90}{Test-Unseen}} & \cite{zhang2016zero} (Acc) & - & - & - & - & - & - & - & - & - & - & 1.37 & 5.69 & 64.35 & 95.83 & 1.12 & 4.00 & 91.52 & 2.31 & 95.41 & 14.91 & - & - & 37.65\\
& ZS-Y+NN & - & - & - & - & - & - & - & - & - & - & \bf{5.24} & 0.32 & \bf{47.60} & \bf{23.53} & \bf{2.69} & \bf{1.13} & \bf{29.97} & \bf{0.49} & \bf{35.70} & \bf{24.15} & - & - & \bf{17.08}\\
& ZS-Y+\cite{zhang2016zero} & - & - & - & - & - & - & - & - & - & - & 0.20 & 0.17 & 11.19 & 18.30 & 0.15 & 1.22 & 11.73 & 0.07 & 3.61 & 0.00 & - & - & 4.66\\
& YOLOv2+\cite{zhang2016zero} & - & - & - & - & - & - & - & - & - & - & 0.47 & \bf{1.05} & 10.96 & 13.59 & 0.44 & 0.54 & 9.64 & 0.43 & 2.33 & 1.97 & - & - & 4.14\\
\hline
\hline
\multirow{2}{*}{\rotatebox{90}{Test-Mix}} & \cite{zhang2016zero} (Acc) & 90.16 & 92.16 & 90.13 & 95.86 & 1.00 & 69.00 & 90.87 & 94.75 & 91.32 & 90.82 & 0.00 & 0.00 & 2.82 & 4.64 & 0.00 & 0.34 & 0.22 & 0.00 & 81.17 & 0.00 &  80.61 & 8.92 & 44.76\\
& ZS-Y+NN & 10.68 & 12.27 & \bf{26.51} & 27.16 & \bf{26.68} & \bf{16.71} & 8.42 & 45.50 & \bf{31.24} & 40.39 & \bf{0.68} & 0.13 & 0.00 & 9.09 & \bf{5.05} & \bf{5.25} & \bf{5.99} & \bf{0.10} & \bf{16.97} & \bf{0.00} & \bf{24.56} & 4.33 & \bf{14.44}\\
& ZS-Y+\cite{zhang2016zero} & 11.94 & \bf{17.89} & 9.70 & 25.89 & 14.72 & 8.93 & \bf{11.09} & 42.99 & 18.85 & 38.96 & 0.00 & \bf{36.36} & \bf{12.80} & 11.31 & 0.00 & 2.51 & 0.13 & 0.00 & 6.09 & 0.00 & 20.10 & \bf{6.92} & 13.51\\
& YOLOv2+\cite{zhang2016zero} & \bf{17.66} & 10.12 & 12.90 & \bf{40.28} & 16.16 & 14.00 & 0.44 & \bf{46.62} & 26.68 & \bf{50.05} & 0.00 & 0.00 & 4.81 & \bf{11.57} & 0.00 & 0.00 & 0.00 & 0.00 & 0.02 & 0.00 & 23.5 & 1.64 & 12.57\\
\hline
\multicolumn{3}{c}{}\\
\end{tabular}
\caption{\small Full detection performance on Pascal VOC 10/10 split\label{table:zsr}}
\end{table*}

\subsubsection{Other Losses} We tried other losses such as the focal loss that has been suggested in the literature to improve detection through hard positive/negative mining. However, we found that while seen object detection improved, we saw a larger loss for unseen objects.

\subsection{Semantic Output for Zero Shot Recognition}

{For gZSL setting, the mean accuracy (1) and mAP in seen and unseen classes are also listed. We emphasize the fact that there are two aspects that contribute to the error. First, error in recognition even when bounding boxes are provided. Second, when bounding boxes are not provided and so both bounding box and recognition is required. Note that GZSL accuracy (namely only the recognition task) for PASCAL VOC is notoriously hard \cite{xian2017zero} with poor accuracy. Consequently, we have a poor baseline to start with and in addition must also perform detection. Our goal in this context is to quantify error increase suffered by virtue of detection. In our ablative study we find that detection error is a smaller component relative to Generalized recognition error for PASCAL VOC.}

By leveraging ZS-YOLO's semantic prediction, ZS-YOLO + NN reaches the highest AP on almost every class in Test-Unsee, and most classes in Test-Mix. Using the same classifier, ZS-YOLO + \cite{zhang2016zero} reaches higher mAP than YOLOv2. This could be attributed to the ability of detecting unseen objects of ZS-YOLO. In the Test-Mix, ZS-YOLO outperforms YOLOv2 on all unseen classes except for cat. It is worth noting that YOLOv2 + \cite{zhang2016zero} achieves 17.66 on bike but 0 on motorbike, suggesting YOLOv2 suppresses unseen objects even they have visually similar seen classes.

{First from Table.~\ref{table:zsr} note that the GZSL error for \cite{zhang2016zero} when bounding boxes are provided is already quite low. While this is not state-of-art, performance on aPY (which is the recognition task on PASCAL VOC) is not much larger with other methods. Our purpose here is less about choosing the best ZSR model and more about detection of unseen objects and so we did not consider other approaches here. Next we observe that both ZS-Y + NN and ZS-Y + \cite{zhang2016zero} achieves much higher mAP on unseen classes. The unseen mAP for ZS-Y + \cite{zhang2015zero} is 6.92\%, which is better than 4.33\% from ZS-Y + NN. However, we notice that ZS-Y + NN outperforms on 7 unseen classes while ZS-Y + \cite{zhang2016zero} only reaches much higher AP on boat and car. Therefore the attribute prediction in ZS-Y is quite accurate and it outperforms cascaded ZSR. ZS-Y + NN outperforms YOLOv2 + \cite{zhang2016zero} on both seen and unseen classes.}
 The error of full zero shot detection comes from the combination of erroneous localization plus misclassification, as the accuracies from \cite{zhang2016zero} are very low for some classes especially in gZSL setting, even though it has no localization error. The results clearly show that our ZS-YOLO can be combined with zero shot recognition methods easily to construct a full zero shot detector and outperforms the original YOLOv2 in the sense of detecting unseen objects. 
 
 {\it Discussion:} We focused on detection aspect of GZSD. In this context, we adopted a simple nearest neighbour model for classification. Nevertheless, we agree that the performance of GZSR classifiers also contributes to the final GZSD performance. Specifically, a key challenge in GZSR is the gap between the semantic domain and 
 the visual domain. There already exists various efforts to bridge such a gap, e.g. \cite{Verma_2018_CVPR,xian2018zero,Chao2016AnES}. Recently, we have found that a promising solution to mitigate the divergence between semantic and visual domains might be to learn a low-dimension 'visually semantic' embedding~\cite{zhu2018generalized, zhu2019learning} that quantifies existence of a prototypical part-type in the presented instance of seen classes. As a possible future direction toward better GZSD performances, one could combine the proposed ZS-YOLO detector with the more powerful ZSR classifiers described above instead of the nearest neighbor classifier adopted in this paper.

\section{Conclusion}
We proposed a novel Zero-Shot YOLO method for unseen object detection that retains the principle advantages of YOLOv2~\cite{redmon2016yolo9000}'s efficiency and performance for seen object detection and extends it non-trivially for unseen object detection. While YOLOv2 is a state-of-art object detection algorithm, its effectiveness hinges on having access to fully annotated datasets, which is unrealistic to expect as as we move towards large-scale object detection. YOLOv2's effectiveness in learning sharp visual features for accurately detecting objects that were seen during training negates its advantages for unseen object detection. To overcome these drawbacks, we propose to build upon YOLOv2's network architecture through seamless fusion of semantic information with the visual domain so that, semantically similar object attributes are also reflected in the learned visual features. Empirically we demonstrated improved performance on two datasets.


%



\section*{Acknowledgment}
The last author would like to thank Dr. Ziming Zhang for initial helpful discussions on zero shot detection and Dr. Tolga Bolubasi for preliminary paper draft and discussions on experiments. The authors would like to thank the Associate Editor and the reviewers for their constructive comments. This work was supported by the Office of Naval Research Grant N0014-18-1-2257, NGA-NURI HM1582-09-1-0037 and the U.S. Department of Homeland Security, Science and Technology Directorate, Office of University Programs, under Grant 2013-ST-061-ED0001.

\ifCLASSOPTIONcaptionsoff
  \newpage
\fi



%



\bibliographystyle{IEEEtran}
\bibliography{ZSD}

%

\begin{IEEEbiography}[{\includegraphics[width=1in,height=1.25in,clip,keepaspectratio]{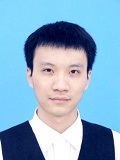}}]{Pengkai Zhu}
Pengkai Zhu is a PhD candidate in the Department of Electrical and Computer Engineering at Boston University, Boston, USA. He received the B.Eng(2014) and M.S.(2016) from Harbin Institute of Technology, Harbin, China. His research interests include computer vision and machine learning, in particular the classification and detection methods under limited resources.
\end{IEEEbiography}

\begin{IEEEbiography}[{\includegraphics[width=1in,height=1.25in,clip,keepaspectratio]{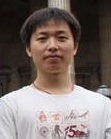}}]{Hanxiao Wang}
Hanxiao Wang is a postdoctoral researcher with the Department of Electrical and Computer Engineering at Boston University. 
He received his B.Eng from Beijing University of Posts and Telecommunications (2013) and PhD from Queen Mary University of London (2017). 
His research interests include computer vision and machine (deep) learning, specially zero-shot learning and video surveillance applications such as cross camera tracking and person re-identification.
\end{IEEEbiography}

\begin{IEEEbiography}[{\includegraphics[width=1in,height=1.25in,clip,keepaspectratio]{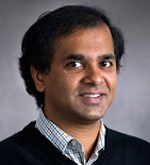}}]{Venkatesh Saligrama}
Venkatesh Saligrama is a faculty member in the Department of Electrical and Computer Engineering and Department of Computer Science (by courtesy) at Boston University. He holds a PhD from MIT. His research interests are in Machine Learning and its applications. He has edited a book on Networked Sensing, Information and Control. He has served as an Associate Editor for IEEE Transactions on Information Theory, IEEE Transactions on Signal Processing and edited special issues and books on Networked Sensing, Detection and Estimation. He is currently serving as the Vice-Chair of Big Data Special Interest Group for IEEE SPS society. He is an IEEE Fellow and recipient of several awards including the Presidential Early Career Award (PECASE), ONR Young Investigator Award, the NSF Career Award and a NIPS 2014 workshop best student paper award on Analysis of Ranking Data. More information about his work is available at http://sites.bu.edu/data.
\end{IEEEbiography}







\end{document}